%% file: acl_latex.tex
\documentclass[11pt]{article}
\usepackage[preprint]{acl}
\usepackage{times}
\usepackage{latexsym}
\usepackage[T1]{fontenc}
\usepackage[utf8]{inputenc}
\usepackage{microtype}
\usepackage{inconsolata}
\usepackage{graphicx}
\usepackage{enumitem}
\usepackage{booktabs}
\usepackage{amssymb}        % 提供 \checkmark 符号
\usepackage{textcomp}       % 提供 \texttimes 符号
\usepackage[table]{xcolor}  % 表格颜色支持
\usepackage{array}          % 增强表格功能
\usepackage{multirow}

\definecolor{lightgray}{gray}{0.92}
\usepackage{booktabs}       % Professional-quality tables
\usepackage{array}          % Advanced column formatting
\usepackage{ragged2e}       % Better text justification
\usepackage{caption}        % Enhanced caption styling
\usepackage{multirow}       % For complex cell merging (if needed later)
\usepackage{xcolor}         % For subtle coloring
\usepackage{tabularx}       % Adjustable width columns
\usepackage{makecell}       % Manual line breaks in cells
\renewcommand{\arraystretch}{1.3} % Increase row height
\definecolor{headerblue}{RGB}{230, 240, 255}

\usepackage{booktabs}
\usepackage{multirow}
\usepackage{pifont}
\usepackage{xcolor}
\usepackage{tabularx}
\usepackage{float}
\usepackage{geometry}
\usepackage{tcolorbox}
\definecolor{boxback}{RGB}{245, 245, 245} % 浅灰背景
\definecolor{boxframe}{RGB}{100, 100, 100} % 深灰边框

\usepackage{tcolorbox}
\tcbuselibrary{skins, breakable}
\usepackage{booktabs} % 用于专业表格线
\usepackage{pifont}   % 用于对勾叉号
\usepackage{enumitem} % 用于紧凑列表

\newcommand{\cmark}{\textcolor{green!60!black}{\ding{51}}}%
\newcommand{\xmark}{\textcolor{red!60!black}{\ding{55}}}%

\newtcolorbox{casestudybox}[2][]{%
    enhanced,
    colframe=gray!60!black,   % 边框颜色（深灰）
    colbacktitle=gray!70!black, % 标题背景色
    coltitle=white,           % 标题文字颜色
    colback=white,            % 内容背景色
    fonttitle=\bfseries\large,
    arc=1mm, boxrule=0.5mm,   % 圆角和边框粗细
    title={#2},               % 标题内容
    #1                        % 可选参数（如 label）
}

\newcommand{\caseheader}[1]{%
    \vspace{0.8em}\noindent{\large\textbf{\textcolor{gray!80!black}{#1}}}\par\vspace{0.2em}\hrule height 0.8pt \vspace{0.5em}%
}
\usepackage[ruled,vlined,linesnumbered]{algorithm2e}
\SetKwInput{KwInput}{Input}
\SetKwInput{KwOutput}{Output}
\SetKw{KwParallel}{parallel}
\SetKw{KwTrigger}{trigger}
\SetKwComment{Comment}{\# }{}
\usepackage{amsmath}
\title{Table-as-Search: Formulate Long-Horizon Agentic Information Seeking\\as Table Completion}

\author{
  Tian Lan, Felix Henry, Bin Zhu, Qianghuai Jia, Junyang Ren \\
  \textbf{Qihang Pu}, \textbf{Haijun Li}, \textbf{Longyue Wang}\thanks{\ \ Corresponding author: wanglongyue.wly@alibaba-inc.com}, \textbf{Zhao Xu}, \textbf{Weihua Luo} \\
  Alibaba International Digital Commerce \\
}

\begin{document}
\maketitle

\begin{abstract}
Current Information Seeking (InfoSeeking) agents struggle to maintain focus and coherence during long-horizon exploration, as tracking search states, including planning procedure and massive search results, within one plain-text context is inherently fragile.
To address this, we introduce \textbf{Table-as-Search (TaS)}, a structured planning framework that reformulates the InfoSeeking task as a Table Completion task.
TaS maps each query into a structured table schema maintained in an external database, where rows represent search candidates and columns denote constraints or required information.
This table precisely manages the search states: filled cells strictly record the history and search results, while empty cells serve as an explicit search plan.
Crucially, TaS unifies three distinct InfoSeeking tasks: Deep Search, Wide Search, and the challenging DeepWide Search.
Extensive experiments demonstrate that TaS significantly outperforms numerous state-of-the-art baselines across three kinds of benchmarks, including multi-agent framework and commercial systems.
Furthermore, our analysis validates the TaS's superior robustness in long-horizon InfoSeeking, alongside its efficiency, scalability and flexibility.
Code and datasets are publicly released at \url{https://github.com/AIDC-AI/Marco-Search-Agent}.
\end{abstract}

\section{Introduction}

%%%%% TODO: 给一个实际的 case，说明限制是什么/candidate 是什么/information 是什么
\begin{figure*}[htbp]
  \centering
  \includegraphics[width=\linewidth]{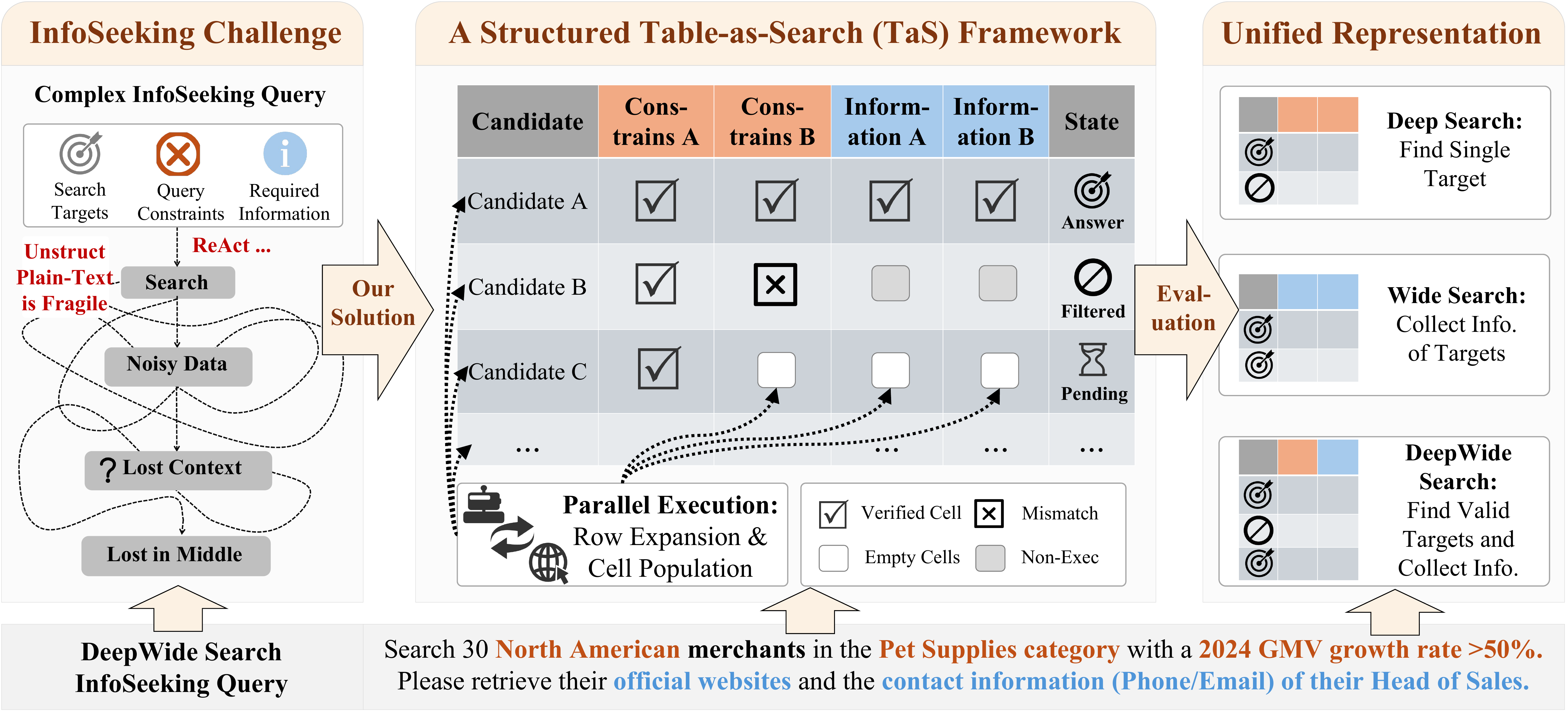}
  \caption{The overview of TaS Framework. \textbf{Left:} Unstructured planning (e.g., ReAct) is fragile and prone to massive context. \textbf{Center:} TaS reformulates InfoSeeking as Table Completion via row expansion and cell population. \textbf{Right:} TaS provides a unified representation for conducting Deep Search, Wide Search and DeepWide Search.}
  \label{fig:overview}
  \vspace{-10pt}
\end{figure*}

Information retrieval is undergoing a paradigm shift from simple fact retrieval to complex long-horizon Agentic InfoSeeking~\cite{deepagent,team2025tongyi,li2025websailornavigatingsuperhumanreasoning,yao2022react}.
It necessitates agents to navigate massive web environments and synthesize answers through multi-step reasoning~\cite{deepagent,team2025tongyi,li2025websailornavigatingsuperhumanreasoning}. 
Mastering this capability is central to next-generation Deep Research Systems~\cite{google2025deepresearch,team2025tongyi}.

%Large Language Model (LLM)-based search agents have emerged as the dominant solution for this task~\cite{team2025tongyi,miromind2025mirothinker}.
%However, the dominant paradigms, e.g., ReAct~\cite{yao2022react},

While Large Language Model (LLM)-based agents have emerged as the dominant solution for this task~\cite{team2025tongyi,miromind2025mirothinker}, current paradigms, such as ReAct~\cite{yao2022react}, rely heavily on unstructured plain text to manage the search states, including planning procedure and massive search results, which is inherently fragile.
Although recent advancements in context management~\cite{wu2025resum,deepagent} and procedural planning~\cite{prasad2024adaptasneededdecompositionplanning,yu2025recodeunifyplanaction} attempt to mitigate this overhead, they still burden the finite unstructured agent context with tracking massive search states of long-horizon InfoSeeking.
Consequently, as the horizon expands, these methods expose agents to the "lost in the middle"~\cite{zhang2024chain} phenomenon, leading to error propagation and ineffective exploration~\cite{chen2025iterresearchrethinkinglonghorizonagents,tao2025webleaper}. 
For instance, tracking thousands of search results and corresponding planning process in WideSearch~\cite{wong2025widesearchbenchmarkingagenticbroad} within a single plain-text trajectory inevitably leads to severe hallucinations and loss of state fidelity.

To address this, we introduce \textbf{Table-as-Search (TaS)}, a structured planning framework that reformulates the InfoSeeking as a \textbf{Table Completion} task.
As illustrated in Figure~\ref{fig:overview}, rather than treating InfoSeeking as unstructured text generation, TaS explicitly maps the user query into a structured schema where rows represent candidate entities and columns denote specific constraints or required information. 
This table precisely manages the search states: filled cells represent the search history and results, while empty cells serve as pending actions (i.e., explicit search plan).
Moreover, by offloading the massive search results to an external database, TaS alleviates the agent's memory burden, preserving the valuable context window for complex reasoning.
Specifically, we implement TaS via a multi-agent system centered around a shared database table. 
A central planner orchestrates sub-agents to iteratively expand rows for candidate discovery and populate cells for constraints verification or information collection.

TaS provides a unified representation for three distinct long-horizon InfoSeeking paradigms:
(1) Deep Search: precise target filtering~\cite{wei2025browsecomp};
(2) Wide Search: broad information aggregation~\cite{wong2025widesearchbenchmarkingagenticbroad};
and (3) the challenging DeepWide Search~\cite{parallel2025findall}: broad exploration and deep verification.
Extensive experiments demonstrate that TaS significantly outperforms state-of-the-art baselines ~\cite{yao2022react,wong2025widesearchbenchmarkingagenticbroad,zhu2025scalingtesttimecomputellm} across these three kinds of benchmarks.
For example, on benchmarks demanding massive search (WideSearch and DeepWide), 
TaS instantiated with the \texttt{Claude-Sonnet-4 (No Think)} significantly outperforms both the computation-heavy Multi-Agent baseline (\texttt{Claude-Sonnet-4 (Thinking)}) and the commercial Gemini DeepResearch system.
Analysis further highlights TaS's superior robustness as InfoSeeking task complexity increases, alongside its efficiency (higher performance with comparable or lower search volume), scalability (effective test-time scaling), and flexibility (seamless integration of specialized deep search agents).

\input{section/related_work}
\input{section/our_method}
\input{section/ds_ws_main_exp}

\input{section/analysis_case_study}

\section{Conclusion}

In this work, we introduced the Table-as-Search (TaS) framework that reformulates long-horizon agentic InfoSeeking as the Table Completion task. TaS maps user query to structured table schema for precise tracking of search states. Extensive experiments demonstrate that TaS significantly outperforms state-of-the-art baselines across Deep, Wide, and DeepWide Search benchmarks. Furthermore, the framework exhibits superior robustness, efficiency, scalability and flexibility, paving the way for more robust InfoSeeking agents.

\section*{Limitation}

\paragraph{Generalization to Non-Search Tasks.} 
While TaS Framework excels in long-horizon InfoSeeking tasks, its applicability to general-purpose agentic tasks remains unstable. The structured tabular schema, optimized for external retrieval and state tracking, may introduce unnecessary rigidity for tasks relying solely on internal knowledge or simple instruction following. This limitation is evidenced by the performance fluctuations observed on non-search GAIA instances (Section~\ref{sec:exp_deep_search}), suggesting that future work should explore adaptive mechanisms to dynamically toggle between structured planning of TaS framework and flexible free-form reasoning based on task demands.

\paragraph{Relationship with Model Optimization.}
It is important to clarify that our contribution is architectural, orthogonal to recent advancements in model training or Agentic Reinforcement Learning (RL)~\cite{li2025websailornavigatingsuperhumanreasoning,miromind2025mirothinker,tao2025webleaper}.
In this work, we do not perform specific fine-tuning for the TaS framework.
However, our ablation studies (Section~\ref{sec:exp_wide_search}) reveal a promising synergy: existing training-based search agents (e.g., WebSailor~\cite{li2025websailornavigatingsuperhumanreasoning}, MiroThinker~\cite{miromind2025mirothinker}) can be seamlessly integrated as Sub-Agents within TaS, boosting execution performance without architectural changes.
This suggests that the Sub-Agents of TaS is plug-and-play compatible with the best open-source models.
Consequently, the critical avenue for future work lies in optimizing the Planner Model.
Developing specialized planners could further mitigate the dependency on proprietary models and fully unlock the potential of the TaS framework.

\paragraph{Dependency on Strong Planner.}
TaS's performance is currently bounded by the reasoning capability of the central Planner Main-Agent. 
As indicated by the ablation study (Section~\ref{sec:analysis_robustness}), while the execution layer (Sub-Agents) can be effectively offloaded to smaller, cost-efficient models without performance loss, the planning layer remains sensitive to model capacity. Downgrading the Planner to weaker models leads to significant performance degradation. 
Our future work will focus on optimizing the Planner—potentially through Agentic RL~\cite{li2025websailornavigatingsuperhumanreasoning,miromind2025mirothinker}.

\paragraph{Distinction from Context Optimization.}
Our core contribution lies in structured planning to enhance search precision, rather than merely mitigating context overflow via compression.
Consequently, recent context optimization strategies (e.g., summarization or folding) are \textbf{orthogonal} to our framework: TaS can also seamlessly incorporate them to further minimize token usage.
However, distinct from these lossy compression methods, TaS offers a unique advantage by offloading critical search states to a structured external database. 
This inherently releases the agent's valuable context window for \textbf{complex reasoning} rather than passive information storage.
Given this fundamental architectural distinction, comparing TaS against pure context compression baselines is unnecessary for validating the efficacy of structured planning.

\paragraph{Evaluation Scalability on DeepWide Search.} A primary limitation of our curated DeepWide Search benchmark lies in the reliance on human evaluation. Unlike closed-domain tasks, DeepWide Search is inherently open-ended, rendering the construction of an exhaustive ground-truth universe computationally infeasible. 
To ensure manageable annotation costs, we explicitly constrain the retrieval target to a fixed quantity for each query (e.g., 30 candidates, as illustrated in Figure~\ref{box:deepwide_case}).
Consequently, accurate assessment currently necessitates human verification to validate whether retrieved candidates strictly satisfy complex constraints. To mitigate the prohibitive cost of annotation and improve efficiency, we implement a dynamic ground-truth maintenance strategy. Specifically, we construct a growing reference dataset by taking the union of verified correct matches (and maintaining an exclusion list for known false positives) across all evaluated systems and human annotation~\cite{parallel2025findall}. While this iteratively updates the ground truth to facilitate partial automation, the dependence on human-in-the-loop verification remains a constraint for large-scale reproducibility.

\bibliography{custom}

\appendix

\input{section/appendix}

\end{document}

%% file: section/related_work.tex
\section{Related Work}

\paragraph{Agentic Information Seeking.}
Recent research categorizes agentic information seeking into three paradigms~\cite{lan2025deepwidesearchbenchmarkingdepthwidth}: Deep Search (multi-step reasoning for single targets)~\cite{mialon2023gaiabenchmarkgeneralai,wei2025browsecomp,zhou2025browsecompzhbenchmarkingwebbrowsing}, Wide Search (broad aggregation across extensive sources)~\cite{wong2025widesearchbenchmarkingagenticbroad,he2025pasallmagentcomprehensive}, and the hybrid DeepWide Search~\cite{parallel2025findall}. While benchmarks exist for the former two (e.g., BrowseComp~\cite{wei2025browsecomp}, WideSearch~\cite{wong2025widesearchbenchmarkingagenticbroad}), the community lacks public high-quality evaluations for DeepWide InfoSeeking. Addressing this gap, we curate a challenging E-commerce Business Development (BD) benchmark, explicitly designed to stress-test agents in real-world DeepWide InfoSeeking.

\paragraph{Agent Frameworks.}
The ReAct paradigm~\cite{yao2022react,liu2025budget} serves as the cornerstone of current agentic systems. 
While recent works have improved ReAct via procedural planning, like Routine~\cite{zeng2025routinestructuralplanningframework}, ADaPT~\cite{prasad2024adaptasneededdecompositionplanning}, ReCode~\cite{yu2025recodeunifyplanaction} and ReCAP~\cite{zhang2025recap}. 
However, these methods still remains bound by unstructured plain-text planning, facing the same problem of ReAct in long-horizon InfoSeeking.
Justified by this shared limitation, we employ the state-of-the-art Multi-Agent ReAct framework~\cite{wong2025widesearchbenchmarkingagenticbroad,kim2025sciencescalingagentsystems} as the representative baseline for these unstructured approaches.
In contrast, TaS is orthogonal to these methods, introducing a data-centric structure to manage massive search states.

\paragraph{Context Management.}
To mitigate context overflow, recent approaches employ strategies like context summarization~\cite{wu2025resum}, folding~\cite{ye2025agentfold} or multi-agent context isolation~\cite{wong2025widesearchbenchmarkingagenticbroad}. 
However, they still suffer from lossy compression and the imprecise unstructured recording of search states.
In contrast, TaS is orthogonal to these strategies; rather than compressing text, it imposes a structured schema on the search process. 
%For example, as detailed in Appendix~\ref{subsec.implementation_details}, both baselines and TaS leverage the context summarization to reduce inference cost, while TaS performs better.
Crucially, while TaS can seamlessly incorporate these strategies (as demonstrated in Section~\ref{subsec.implementation_details}), its distinct advantage lies in offloading massive search results to a structured external database for on-demand access, reserving agent's reasoning capacity for complex decision-making rather than passive information storage.

%% file: section/our_method.tex
%\section{Task Formulation}
% 该章节用于形式化定义 IS 任务和table-center 的 framework 任务。
%\subsection{Formulation of IS and IS Agent}
% 定义任务，可以参考图片中的写法，但是不要照抄！请你结合我们的论文的主要 motivation 来介绍 task formation
%\subsection{Formulation of Table-center Framework}
% 定义我们的框架如何解决这个任务的：如 figure 1 所示，一个IS query通常包含三个元素：（1）search target；（2）constratints；（3）related information need to be searched. xxxxx 我们怎么拆解 query 的，怎么样通过表格的方式就能解决这个任务
% 以及怎么样把表格框架解决 IS 问题的方式推广到 deep search，wide search 和 deepwides search

\section{Task Formulation}

\subsection{Problem Definition}
\label{sec:is_formulation}

Formally, an InfoSeeking task is defined as a tuple $\mathcal{T} = \langle q, \mathcal{W} \rangle$, where an agent interacts with the web environment $\mathcal{W}$ to fulfill a complex query $q$. The interaction unfolds over $T$ steps, generating a trajectory (history) $\tau_T = (o_1, r_1, a_1, \dots, o_T, r_T, a_T)$, where $o_t$, $r_t$ and $a_t$ denote observations, chain-of-thoughts and actions, respectively~\cite{fang2025webevolverenhancingwebagent}.
Standard paradigms (e.g., ReAct) model the agent's policy $\pi$ as generating the next action conditioned on the entire unstructured history $\tau_t$: 
$r_{t+1},a_{t+1} \sim \pi(\cdot \mid q,\tau_t)$.
Critically, as the horizon $t$ extend, the relevant information density in $\tau_t$ dilutes, causing the "lost-in-the-middle" phenomenon~\cite{chen2025iterresearchrethinkinglonghorizonagents}. The agent must implicitly perform information extraction and state tracking simultaneously within a single forward pass. This challenges agents to propose plans for effective exploration in the search space.

\subsection{Table-as-Search (TaS) Framework}
\label{sec:table_formulation}
To resolve this, we reformulate the InfoSeeking task as a Table Completion problem for precise search state management.
\paragraph{Structured Schema Definition.} Instead of operating on free-form text, we map the query $q$ into a structured schema $\mathcal{S}$: $\phi(q) \to \mathcal{S}$. The schema is defined as a tuple of attribute sets: $\mathcal{S} = \langle \mathcal{K}, \mathcal{C}, \mathcal{I} \rangle$. $\mathcal{K}$ uniquely represents the key candidates, $\mathcal{C}$ denotes the Constraint Set, and $\mathcal{I}$ denotes the Information Set (information to be collected). This formulation generalizes to distinct InfoSeeking paradigms by simply varying the set configurations.

\paragraph{Search as Table Completion.} As shown in Figure~\ref{fig:overview}, we maintain the long-horizon InfoSeeking as a table $T_t$, where rows correspond to discovered or potential candidates $e \in \mathcal{E}$ and columns correspond to the schema $\mathcal{S}$. Let $T_t[i, j]$ denote the cell for the $i$-th candidate and $j$-th attribute. The cell takes values from $\mathcal{V} \cup \{\emptyset, \text{N/A}\}$, where $\emptyset$ represents a "pending" state and \text{N/A} denotes the information that do not need to retrieve.
%Crucially, we map the complex InfoSeeking to the table completion (i.e., $\forall i,j, T_t[i,j] \neq \emptyset$), a simple bounded completion task.
Under this formulation, the policy $\pi$ is conditioned on a structured table and trajectory: $r_{t+1},a_{t+1} \sim \pi(\cdot \mid q, \tau_t,T_t)$. Once $T_t$ is fully populated, the complex query $q$ can be answered by referring the evidence in $T_t$.

\paragraph{Unified View of InfoSeeking.} This tabular formulation provides a unified representations of three distinct InfoSeeking paradigms:
(1) Deep Search (Precise Filtering): The objective is to identify a unique candidate row that strictly satisfies all constraints ($|\mathcal{C}|>0$), often involving complex multi-hop verification to filter out false positives;
(2) Wide Search (Broad Aggregation): The primary goal is to gather required information ($|\mathcal{I}| > 0$) for a massive candidates, typically under minimal constraints~\cite{wong2025widesearchbenchmarkingagenticbroad};
(3) DeepWide Search (Hybird): A complex hybrid scenario requiring the maximization of candidate discovery subject to strict constraint satisfaction, followed by dense information collection ($|\mathcal{C}|>0,|\mathcal{I}|>0$).

\section{Implementation of TaS Framework}
\label{sec:method_system}

We instantiate the TaS framework as a multi-agent system centered around a shared, structured database table. 
%The Main Agent functions as the planner, managing the table state and calling Sub-Agent to search candidates or populate empty cells in $T$.
As outlined in Algorithm~\ref{alg:table_execution} and Figure~\ref{fig:show_case_control_tas}, the execution follows a three-phase process.

\begin{algorithm}[t]
\RestyleAlgo{boxruled} % 强制改为带框风格
\caption{\textbf{Multi-Agent System of TaS}}
\label{alg:table_execution}
\footnotesize % 保持小字号
\DontPrintSemicolon

% === 关键优化：减小缩进宽度，腾出右侧空间 ===
\SetInd{0.8em}{0.8em} 

\SetKwInOut{Input}{Input}
\SetKwInOut{Output}{Output}

\Input{Query $Q$, MaxSteps $T_{max}$, Timeout $\tau$}
\Output{Final synthesized answer $A$}

\tcp{Phase 1: Table Initialization}
\tcp{Define Key Cands./Cons./Info columns}
$S \leftarrow \text{MainAgent}.\text{ConstructSchema}(Q)$; 

$Table \leftarrow \text{Initialize}(S)$; $State \leftarrow \text{Pending}$\; $\mathcal{H}_T \leftarrow \{\}$\;

\tcp{Phase 2: Dynamic Orchestration}
\While{$State \neq \text{Done} \land \neg \text{Limits}(T_{max}, \tau)$}{
    $Plan \leftarrow \text{Main}.\text{FormulateStrategy}(Table, Q)$\;
    
    \If{$Plan.\text{action} == \text{ExpandRows}$}{
        \tcp{Case: No enough valid candidates}
        % === 优化：移除强制换行，让它自然伸展 ===
        $\{q\}_{i=0}^n \leftarrow \text{MakeQuery}(Table.\text{ConsCols})$\;
        \ForEach{$q_{i}$ \textbf{in parallel}}{
            $Cands \leftarrow \text{SubAgent}.\text{DeepSearch}(q_i)$\;
            $Table.\text{AppendRows}(Cands)$\;
        }
    }
    
    \If{$Plan.\text{action} == \text{PopulateCells}$}{
        \tcp{Row-Level Parallel Execution}
        % 修改点1：获取不完整的行，而不是零散的单元格
        $Rows \leftarrow Table.\text{GetIncompleteRows}()$\;
        
        % 修改点2：按行并行
        \ForEach{$R_{i} \in Rows$ \textbf{in parallel}}{
            % 修改点3：任务是填补该行所有缺失的列
            $q_{i} \leftarrow \text{MakeQuery}(R_i, Table.\text{EmptyInfoCols})$\;
            % SubAgent 处理整行逻辑，保持上下文连贯
            $Res_{i} \leftarrow \text{SubAgent}.\text{DeepSearch}(q_{i})$\;
            $Table.\text{UpdateRow}(R_i, Res_{i})$\;
        }
    }
    $State \leftarrow \text{Main}.\text{CheckSaturation}(Table)$\;
    $\text{Main}.\text{Update}(\mathcal{H}_T)$\;
}
\tcp{Phase 3: Answer Synthesis}
\Return $\text{Main}.\text{Synthesize}(Table, Q)$
\end{algorithm}

\paragraph{Table Initialization.}
The Planner parses the user query $q$ and initialize the table structure in the database (\texttt{ConstructSchema}).

\paragraph{Dynamic Orchestration.}
In the main loop (Lines 4-18), the Planner Main-Agent dynamically selects the action: (1) \textbf{Row Expansion (Lines 6-10)}: For example, if the table lacks candidates, or if current candidates fail to satisfy query constraints, it formulates $n$ diverse search strategies using the constraints (Line 7). These strategies are orchestrated to Sub-Agents in parallel to perform broad searches, aiming to discover new candidates;
(2) \textbf{Cell Population (Lines 11-16)}: Conversely, if candidates are sufficient but their information is incomplete, the system transitions to this mode. Leveraging the independence of candidates, the Main Agent dispatches Sub-Agents in parallel to populate cells for each candidate. 
Notably, TaS allows for high flexibility: Since Sub-Agents inherently align with the recent specialized deep search models~\cite{miromind2025mirothinker,li2025websailornavigatingsuperhumanreasoning}, TaS can seamlessly integrate advanced off-the-shelf search agents as sub-agents.
Both Main-Agent and Sub-Agent manipulate table (\texttt{AppendRow} in Line 10 and \texttt{UpdateRow} in Line 16) via database interface. More details are in Appendix~\ref{appendix:exp_setup}.

\paragraph{Answer Synthesis.}
Upon detecting a saturated table state (or timeout), the Planner retrieves the structured evidence from the database to synthesize the final response $A$. For example, for Deep Search, the planner utilizes the filled table to cross-verify constraints for a precise conclusion; conversely, for Wide Search and DeepWide Search, it directly executes SQL queries to export the verified candidates.

%\paragraph{Advantages of TaS Framework.}
% 介绍 parallism和 human-readbility（本质上是对 trajectory 的高度准确的压缩）的优势？

%% file: section/ds_ws_main_exp.tex
% and metrics
%\subsection{Benchmarks}
% 1. 我们测试了三类information seeking的 benchmark：
%（1）Deep search 测试了 GAIA-text-only和 BrowseComp-ZH；评估指标采用 LLM-as-a-Judge 汇报的 Accuracy；
%（2）Wide Search 测试了widesearch benchmark，由于实验开销的限制，实验我们主要汇报 Pass@1 的结果，pass@4 的分析结果我们针对 gemini-2.5-flash 模型单独分析； 评估指标采用 Column-F1（实体查找准确率），Row-F1（行级别准确率），Item-F1（单元格级别准确率） 和 Success rate（全表准确率）。
%（3）deepwidesearch：我们针对真实业务应用场景，构建了电商领域招商 BD 场景的 20 条极为复杂的深宽搜索数据样本，并通过大量模型采样和人工标注构建标准答案（开销极大）。测试指标为 Column-F1（实体查找准确率）和Item-F1(线索查找准确率)。测试结果采用人工标注完成以确保准确性。

% base models, training-based search agents
%\subsection{Baselines}
% 我们测试了两大类 baseline：
% （1）foundation model with tools：先进的开源或者闭源的foundation models 和工具：包括 deepseek-r1, openai o3, claude-sonnet-4, gemini-2.5-pro, gemini-2.5-flash, qwen3-max, qwen-3-235b-a22b, qwen3-next-80b, qwen3-30b等；
% （2）training-based search agent with tools：经过大规模 agentic RL 训练的模型和工具调用，如websailor-7b/32b, tongyi deepresearch 32B, mirothinker-v1.0-8/30/70B.
% 其中所有模型工具均为 google search 和 webpage visit。

\section{Experimental Setup}

\subsection{Benchmarks and Metrics}
To rigorously evaluate TaS across distinct long-horizon agentic infoseeking, we employ three categories of benchmarks: 
(1) \textbf{Deep Search}:
We utilize GAIA (text-only)~\cite{mialon2023gaiabenchmarkgeneralai} and BrowseComp-ZH~\cite{zhou2025browsecompzhbenchmarkingwebbrowsing} to assess multi-step reasoning and precise filtering capabilities. Performance is measured by Accuracy, evaluated via standard LLM-as-a-Judge protocols~\cite{zhou2025browsecompzhbenchmarkingwebbrowsing}; 
(2) \textbf{Wide Search}: We employ the WideSearch benchmark to evaluate broad information aggregation~\cite{wong2025widesearchbenchmarkingagenticbroad}. To reduce the randomness, we report the stable Avg@4 metrics of Column-F1 (Candidate Acc.), Row-F1 (Row-level Acc.), Item-F1 (Cell-level Acc) and Success Rate (SR, Table-level Acc.);
(3) \textbf{DeepWide Search}: As existing benchmarks lack scenarios requiring both extensive candidate discovery and deep constratins verification and information collection, we curate a benchmark consisting of 20 challenging long-horizon InfoSeeking queries derived from real-world E-commerce scenarios (e.g., sourcing merchants meeting strict criteria). Given the high cost of expert curation, this dataset size aligns with concurrent studies~\cite{parallel2025findall}. Cases can be found in Figure~\ref{box:deepwide_case}. We employ expert annotation to report Column-F1 and Item-Precision (Information Correctness) due to the open-ended complexity.

\paragraph{Experimental Scale and Cost.} Some may argue for broader benchmark coverage. However, given the prohibitive cost of long-horizon execution (over \$5,000), our setup ensures a representative evaluation while maintaining computational feasibility.

\subsection{Baseline Models and Systems}
We compare TaS against two kinds of baselines: 
(1) \textbf{Agentic Frameworks}: We evaluate standard Single-Agent ReAct (ReAct-SA)~\cite{yao2022react,tao2025webleaper}, Multi-Agent ReAct (ReAct-MA)~\cite{wong2025widesearchbenchmarkingagenticbroad,kim2025sciencescalingagentsystems}, and their compute-scaled variants~\cite{zhu2025scalingtesttimecomputellm}. 
Multi-Agent serves as the state-of-the-art baseline in Wide Search~\cite{wong2025widesearchbenchmarkingagenticbroad} and Deep Search (as evidenced in Table~\ref{tab:deep_search_perf}).
These frameworks are instantiated with diverse foundation models, including GPT-5, Claude-Sonnet-4, Gemini-2.5 series, KIMI-K2~\cite{kimiteam2025kimik2openagentic}, Qwen3 series~\cite{yang2025qwen3technicalreport}, etc.;
(2) \textbf{State-of-the-Art Systems}: We further benchmark against specialized search agents, including commercial systems (Gemini DeepResearch) and models trained by Agentic RL~\cite{miromind2025mirothinker,tao2025webleaper}.

\subsection{Implementation Details~\label{subsec.implementation_details}}

Our experiments are based on the SmolAgent framework~\cite{smolagents} and WideSearch~\cite{wong2025widesearchbenchmarkingagenticbroad}.
All agents utilize two standard tools: Google Search and Webpage Visit~\cite{wong2025widesearchbenchmarkingagenticbroad} to interact with environments.
All training-based search sub-agents are served on a cluster of 8 NVIDIA A100 GPUs. To handle long contexts, we set the maximum context window to 64k tokens.
We integrate webpage and context summarization strategies for reducing cost~\cite{miromind2025mirothinker,wu2025resum}. 
Full hyperparameters, prompt details and table tool implementation in TaS are provided in Appendix~\ref{appendix:exp_setup}.

\section{Main Results}

This section provide experimental results on three kinds of Agentic InfoSeeking benchmarks: (1) Deep Search  (Section~\ref{sec:exp_deep_search}); (2) Wide Search (Section~\ref{sec:exp_wide_search}) and (3) DeepWide Search (Section~\ref{sec:exp_deepwide_search}).

\subsection{Results on Deep Search Benchmarks\label{sec:exp_deep_search}} 
Table~\ref{tab:deep_search_perf} and Table~\ref{tab:gaia_breakdown} presents the comparative analysis on GAIA and BrowseComp-ZH benchmarks.
\paragraph{TaS Outperforms Unstructured Baselines.} TaS consistently outperforms most Single-Agent and Multi-Agent ReAct baselines across diverse backbone models. Most notably, when instantiated with the cost-efficient Gemini-2.5-Flash, our framework surpasses the Multi-Agent ReAct baseline by a substantial margin of +14.0\% on GAIA (52.4\% vs. 38.4\%), outperforming better counterpart Qwen3-Max. This result confirms that the performance bottleneck in weaker models is often not reasoning capability, but search state management. By maintaining the search state into a structured table, TaS effectively enables smaller models to perform on par with significantly larger counterparts.
\paragraph{Superiority in InfoSeeking Setup.} We observe a slight regression on GAIA (49.0\% vs. 52.0\%). However, the breakdown in Table~\ref{tab:gaia_breakdown} reveals that this drop is strictly confined to non-search tasks (-18.2\%), where the structured table overhead is unnecessary for simple internal agentic tasks. Crucially, on the search-dependent subset central to our objective, TaS maintains its superiority (+2.5\%).

\begin{table}[h]
\centering
\resizebox{\linewidth}{!}{
% 修改这里：增加了一列 c (总共4列)
\begin{tabular}{l c cc}
\toprule
\textbf{Model / System} & \textbf{Type} & \textbf{GAIA} & \textbf{BC-ZH} \\
\midrule
% 修改这里：跨列数改为 4
\multicolumn{4}{c}{\textbf{Foundation Models with Tools}} \\
\midrule
%DeepSeek-R1 & - & 31.1 & 26.3 \\
%OpenAI o3 & - & - & 58.1 \\
OpenAI Deep Research & - & 67.4 & 42.9 \\
GPT-5 High-Think & - & 76.4 & 63.0 \\
Claude-4-Sonnet (Thinking) & SA & 68.3 & 29.1 \\
Gemini-2.5-Pro & SA & 60.2 & 27.8\\
\midrule
\multicolumn{4}{c}{\textbf{Training-based Search Agents}} \\
\midrule
%WebSailor-7B & - & 37.9 & 9.7 \\
%WebSailor-32B & SA & 53.2 & 25.5 \\
Tongyi DeepResearch (30B) & SA & 70.9 & 46.7 \\
MiroThinker-v1.0-8B & SA & 66.4 & 40.2 \\
MiroThinker-v1.0-30B & SA & 73.5 & 47.8 \\
MiroThinker-v1.0-72B & SA & 81.9 & 55.6 \\
\midrule
\multicolumn{4}{c}{\textbf{Our proposed TaS Framework}} \\
\midrule
% 这里移除了原本在 Deep Search 列全为 "-" 的行
%\rowcolor{lightgray}
%Qwen3-235B-A22B (Ours) & - & 32.4 & 29.9 \\
GPT-5 Medium-Think & SA & 66.0 & 56.5 \\
GPT-5 Medium-Think & MA & 71.8 & 62.9 \\
\rowcolor{lightgray}
GPT-5 Medium-Think (Ours) & MA & \textbf{77.7} & \textbf{63.7} \\
Qwen3-Max & SA & 39.8 & 23.5 \\
%Qwen3-Max (ReAct-SA-MV) & SA-MV & 50.0 & 25.6 \\
Qwen3-Max & MA & \textbf{52.0} & 34.3 \\
\rowcolor{lightgray}
Qwen3-Max (Ours) & MA & 49.0 & \textbf{35.3} \\
Gemini-2.5-Flash & SA & 16.3 & 26.6 \\
%Gemini-2.5-Flash (ReAct-SA-MV) & SA-MV & 25.2 & 12.5 \\
Gemini-2.5-Flash & MA & 38.4 & 28.4 \\
\rowcolor{lightgray}
Gemini-2.5-Flash (Ours) & MA & \textbf{52.4} & \textbf{34.9} \\
\bottomrule
\end{tabular}
}
\caption{Performance Comparison on \textbf{Deep Search Benchmarks}. BC-ZH refers to BrowseComp-ZH.}
\label{tab:deep_search_perf}
%\vspace{-10pt}
\end{table}

\begin{table}[t]
\centering
\resizebox{\linewidth}{!}{
\begin{tabular}{ll cccc}
\toprule
\textbf{Model} & \textbf{Sub-Task Type} & \textbf{Num} & \textbf{ReAct} & \textbf{Ours} & \textbf{$\Delta$} \\
\midrule
% GPT-5 部分
%\multirow{3}{*}{\shortstack[l]{\textbf{GPT-5}\\\textbf{Medium}\\\textbf{Think}}} 
% & Requires Search & 66.25\% & \textbf{71.25\%} & \textcolor{teal}{+5.0\%} \\
% & No Search & \textbf{91.30\%} & 86.96\% & \textcolor{red}{-4.34\%}\\
% \cmidrule(lr){2-5}
% & \textit{Overall} & 71.84\% & \textbf{77.67\%} & \textcolor{teal}{+5.87\%}\\
%\midrule
% Qwen3-Max 部分
\multirow{3}{*}{\shortstack[l]{\textbf{Qwen3}\\\textbf{-Max}}} 
 & Requires Search & 80 & 46.8\% & \textbf{49.4\%} & \textcolor{teal}{+2.5\%} \\
 & No Search & 23 & \textbf{68.2\%} & 50.0\% & \textcolor{red}{-18.2\%} \\
 \cmidrule(lr){2-6}
 & Overall & 103 & \textbf{51.5\%} & 49.5\% & \textcolor{red}{-2.0\%} \\
\midrule
% Gemini-2.5-Flash 部分
\multirow{3}{*}{\shortstack[l]{\textbf{Gemini}\\\textbf{2.5-Flash}}} 
 & Requires Search & 80 & 34.2\% & \textbf{49.4\%} & \textcolor{teal}{+15.2\%} \\
 & No Search & 23 & 55.0\% & \textbf{60.0\%} & \textcolor{teal}{+5.0\%} \\
 \cmidrule(lr){2-6}
 & Overall & 103 & 38.4\% & \textbf{51.5\%} & \textcolor{teal}{+13.1\%} \\
\bottomrule
\end{tabular}
}
% Caption 说明：强调我们在 "Requires Search" 子集上的一致性优势
\caption{Detailed Performance on GAIA. Please refer to Appendix~\ref{appendix:case_study_gaia} for more details.}
\label{tab:gaia_breakdown}
\end{table}

\subsection{Results on Wide Search Benchmark\label{sec:exp_wide_search}} 
Table~\ref{tab:wide_search_full} demonstrates the Avg@4 performance on WideSearch~\cite{wong2025widesearchbenchmarkingagenticbroad}, which is suited to stress-test agents due to its massive search space (Avg. 274.8 table cells per query). Max@4 performance is shown in Table~\ref{tab:wide_search_full_max@4}.

\paragraph{Superiority of TaS Framework.} TaS demonstrates holistic superiority over state-of-the-art baselines. 
%A striking observation is that TaS significantly reduces the dependency on expensive reasoning models. 
As shown in Table~\ref{tab:wide_search_full}, TaS with \texttt{Claude-Sonnet-4 (NoThink)} achieves comparable performance to the ReAct-MA with \texttt{Claude-Sonnet-4 (Thinking)} on Success Rate (3.5\% $\approx$ 3.6\%). Besides, Max@4 Performance in Table~\ref{tab:wide_search_full_max@4} shows that TaS with \texttt{Claude-Sonnet-4 (NoThink)} significantly surpassing ReAct-MA (\texttt{Claude-Sonnet-4} \texttt{(Thinking})) on Success Rate (9.1\% $>$ 6.5\%), exhibiting higher potentional.
Moreover, instantiated with the lightweight \texttt{Gemini-2.5-Flash}, TaS outperforms ReAct-MA baseline running on the much stronger \texttt{Gemini-2.5-Pro} (Success Rate: 2.2\% $>$ 2.0\%). This inversion indicates that in long-horizon tasks, the performance bottleneck shifts from reasoning capability to state management, where TaS's structured planning enables smaller models to rival significantly larger counterparts.

\begin{table}[htbp]
\centering
% 保持紧凑的列间距
\setlength{\tabcolsep}{3pt}
% 保持舒缓的行高
\renewcommand{\arraystretch}{1.15}
\resizebox{\linewidth}{!}{
% 修改这里：增加了一列 c (总共6列)
\begin{tabular}{l c cccc}
\toprule
\multirow{2}{*}{\textbf{Model}} & \textbf{ReAct} & \textbf{SR} & \textbf{Row} & \textbf{Item} & \textbf{Col} \\
& \textbf{Type} & \textbf{Acc} & \textbf{F1} & \textbf{F1} & \textbf{F1} \\
\midrule
% 修改这里：跨列数改为 6
\multicolumn{6}{c}{\textbf{Foundation Models with Tools}} \\
\midrule
Claude-S4 Think & SA & 2.3 & 31.7 & 57.9 & - \\
Claude-S4 Think & MA & 3.6 & 38.5 & 62.2 & - \\
Gemini-2.5-Pro  & SA & 1.5 & 30.0 & 51.0 & -\\
Gemini-2.5-Pro  & MA & 2.0 & 33.5 & 57.4 & -\\
OpenAI o3       & SA & 4.5 & 34.0 & 52.6 & - \\
OpenAI o3       & MA & 5.1 & 37.8 & 57.3 & - \\
KIMI-K2         & SA & 1.1 & 29.7 & 54.4 & - \\
KIMI-K2         & MA & 3.0 & 36.2 & 61.2 & - \\
WebLeaper       & SA & 4.0 & 31.0 & 48.8 & - \\ 
\midrule
% 修改这里：跨列数改为 6
\multicolumn{6}{c}{\textbf{Our proposed TaS Framework}} \\
\midrule
Gemini-2.5-Flash & SA & 2.0 & 26.9 & 49.9 & 62.1\\
Gemini-2.5-Flash & MA & 1.9 & 26.3 & 45.7 & 55.4 \\
\rowcolor{lightgray}
Gemini-2.5-Flash (Ours) & MA & \textbf{2.2} & \textbf{29.1} & \textbf{52.7} & \textbf{66.8} \\
Claude-S4 NoThink & SA & 2.2 & 26.1 & 48.6 & 61.3 \\
Claude-S4 NoThink & MA & 3.2 & 33.7 & 56.6 & 68.0 \\ 
\rowcolor{lightgray}
Claude-S4 NoThink (Ours) & MA & \textbf{3.5} & \textbf{36.7} & \textbf{60.5} & \textbf{74.7} \\
\bottomrule
\end{tabular}
}
\caption{\textbf{Avg@4} Performance on \textbf{WideSearch benchmark}. Claude-S4 denotes \texttt{Claude-Sonnet-4}. Baseline results are copied from \citet{wong2025widesearchbenchmarkingagenticbroad}, where Column-F1 scores are not recorded.}
\label{tab:wide_search_full}
\end{table}

\paragraph{Better Precision-Recall Trade-off.} Typically, expanding the search horizon in precision-recall trade-off, where aggressive exploration introduces noise and hallucinations. However, as shown in Table~\ref{tab:wide_search_detailed}, TaS simultaneously improves both precision and recall performance. Specifically, TaS significantly boosts in Column-Recall (+8.4\%) and Item-Recall (+6.9\%) compared to the ReAct-MA. Crucially, this higher coverage does not come at the cost of precision (e.g. +4.4\% in Item-Precision), validating the table constraints effectively filter out noise during the extensive information gathering.

\begin{table}[htbp]
\centering
\setlength{\tabcolsep}{3pt}
\renewcommand{\arraystretch}{1.15}
\resizebox{0.95\linewidth}{!}{
\begin{tabular}{l c ccc}
\toprule
\textbf{Model} & \textbf{ReAct} & \textbf{Row} & \textbf{Item} & \textbf{Col} \\
\midrule
\multicolumn{5}{c}{\textbf{Precision Performance}} \\
\midrule
Claude-S4 NoThink & SA & 31.0 & 54.6 & 75.5 \\
Claude-S4 NoThink & MA & 37.6 & 63.6 & 78.4 \\ 
\rowcolor{lightgray}
Claude-S4 NoThink (Ours) & MA & \textbf{39.6} & \textbf{68.0} & \textbf{84.6} \\
\midrule
\multicolumn{5}{c}{\textbf{Recall Performance}} \\
\midrule
Claude-S4 NoThink & SA & 23.6 & 44.6 & 56.0 \\
Claude-S4 NoThink & MA & 31.8 & 51.9 & 64.0 \\ 
\rowcolor{lightgray}
Claude-S4 NoThink (Ours) & MA & \textbf{34.2} & \textbf{58.8} & \textbf{72.4} \\
\bottomrule
\end{tabular}
}
\caption{Detailed Avg@4 Precision-Recall Performance of \texttt{Claude-Sonnet-4} on the WideSearch benchmark.}
\label{tab:wide_search_detailed}
\end{table}

\subsection{Results on DeepWide Search Benchmark\label{sec:exp_deepwide_search}}

\begin{table}[htbp]
\centering
\resizebox{0.95\linewidth}{!}{
\begin{tabular}{lccc}
\toprule
\textbf{Models / Systems} & \textbf{ReAct} & \textbf{Col-F1} & \textbf{Item-P} \\
\midrule
Gemini DeepResearch & - &51.2 &58.3 \\
Claude-Sonnet-4 & SA & 39.5 & 35.2 \\
Claude-Sonnet-4 & MA & 39.3 & 44.2 \\ 
\midrule
\multicolumn{4}{c}{\textbf{Our proposed TaS Framework}} \\
\midrule
\rowcolor{lightgray}
Claude-Sonnet-4 (TaS) & MA & \textbf{55.9} & 63.5 \\
\rowcolor{lightgray}
\ \ + 32B Sub-Agent & MA & 52.7 & \textbf{67.7} \\
\bottomrule
\end{tabular}
}
\caption{Performance on \textbf{DeepWide Search Benchmark}. Baselines and TaS use \texttt{Claude-Sonnet-4}.}
\label{tab:deewidesearch_result}
\end{table}

\paragraph{Superior Performance.} On the challenging DeepWide benchmark, TaS demonstrates decisive superiority. As shown in Table~\ref{tab:deewidesearch_result}, it outperforms not only ReAct-MA but also the state-of-the-art Gemini DeepResearch, achieving gains of +4.7\% in Column-F1 and +5.1\% in Item-Precision. This confirms that explicit structured planning provides a critical edge over proprietary black-box systems in complex long-horizon InfoSeeking tasks.

\paragraph{Flexibility and Efficiency.} TaS further proves its architectural scalability by effectively decoupling planning from execution. As shown in the last row of Table~\ref{tab:deewidesearch_result}, replacing the sub-agent with a fine-tuned 32B deep search model yields a promising result: while candidate discovery sees a marginal trade-off (Column-F1: 55.9\% $>$ 52.7\%), the information retrieval precision significantly improves (Item-Precision: 67.7\% $>$ 63.5\%).
This result confirms that high-frequency search actions can be offloaded to cost-effective specialized model to boost precision, making TaS a highly flexible and efficient solution for industrial-scale applications.

\section{Analysis\label{sec:analysis}}
We investigate the underlying mechanisms of TaS through four critical research questions (RQs). Specifically, we examine whether structured planning enhances Robustness in long-horizon InfoSeeking (RQ1) and improves Efficiency beyond simple scaling search volume (RQ2). We further analyze the Test-Time Scaling (RQ3) and Ablation Studies (RQ4) to compare the planner versus the sub-agents. Detailed experimental setup and results are provided in Appendix~\ref{appendix:exp_setup_analysis} and Appendix~\ref{appendix:more_exp_rest}.

\subsection{Robustness on Long-Horizon InfoSeeking~\label{sec:analysis_robustness}}

\begin{figure}[htbp]
    \centering
    \includegraphics[width=0.48\textwidth]{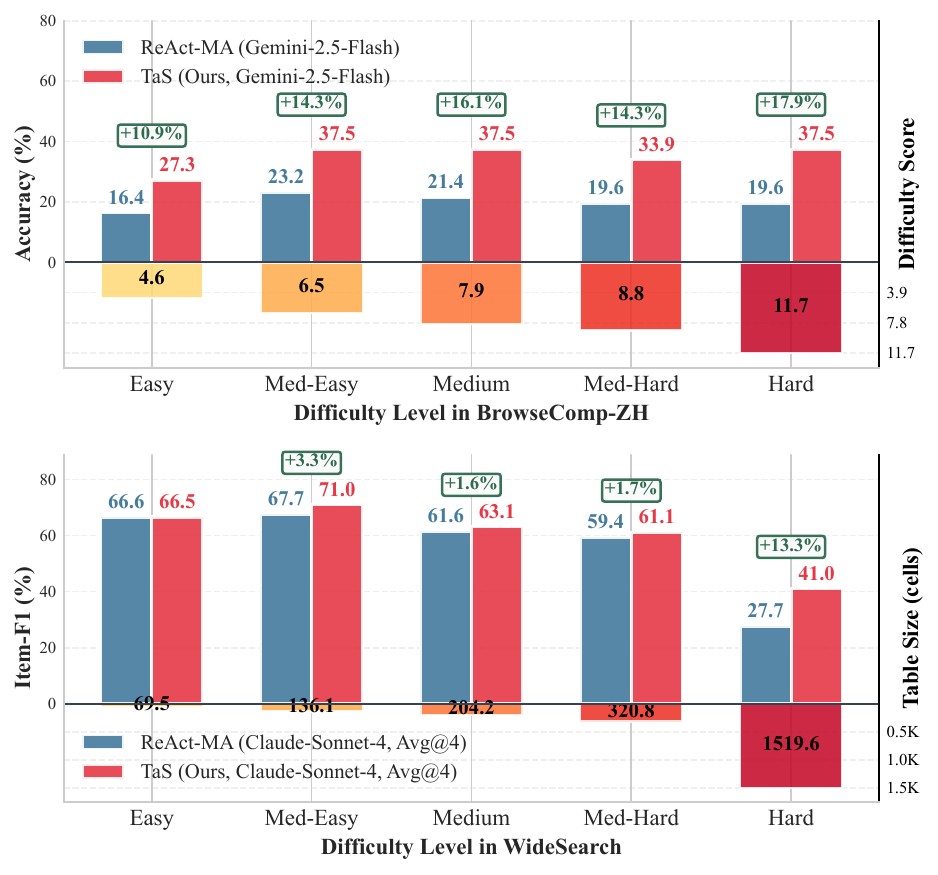}
    \caption{Robustness Analysis on BrowseComp-ZH (Top) and WideSearch (Bottom).}
    \label{fig:robustness_stack}
    %\vspace{-5pt}
\end{figure}

\paragraph{RQ1: Is TaS robust to increasing complexity in long-horizon InfoSeeking?}
We classify instances in benchmarks into five difficulty levels based on distinct complexity metrics: constraint count $|\mathcal{C}|$ (searching complexity) for Deep Search, and table size (interaction horizon) for Wide Search. As visualized in Figure~\ref{fig:robustness_stack}, TaS demonstrates widening superiority as complexity scales:
(1) \textbf{Deep Search} (Top): The performance gap over baselines expands from +14.3\% in Med-Hard to +17.9\% in the Hard instances. Crucially, TaS maintains consistent accuracy levels that match or even exceed those of easier tiers, validating its stability in deep reasoning;
(2) \textbf{Wide Search} (Bottom, \texttt{Claude-Sonnet-4}): The superiority of TaS is highlighted by the drastic expansion of the performance gap from Med-Hard (+1.7\%) to the Hard tier (+13.3\%). This divergence indicates that while baselines experience a complete breakdown ($>$ 30\%), TaS exhibits a much slower rate of decay, effectively tracking search states.

\begin{figure*}[ht]
    \centering
    % width=\textwidth 确保图片撑满整个页面的宽度（即跨两栏）
    % height=6cm 只是为了占位演示，实际使用时可以去掉 height 参数
    \includegraphics[width=\textwidth, height=5cm]{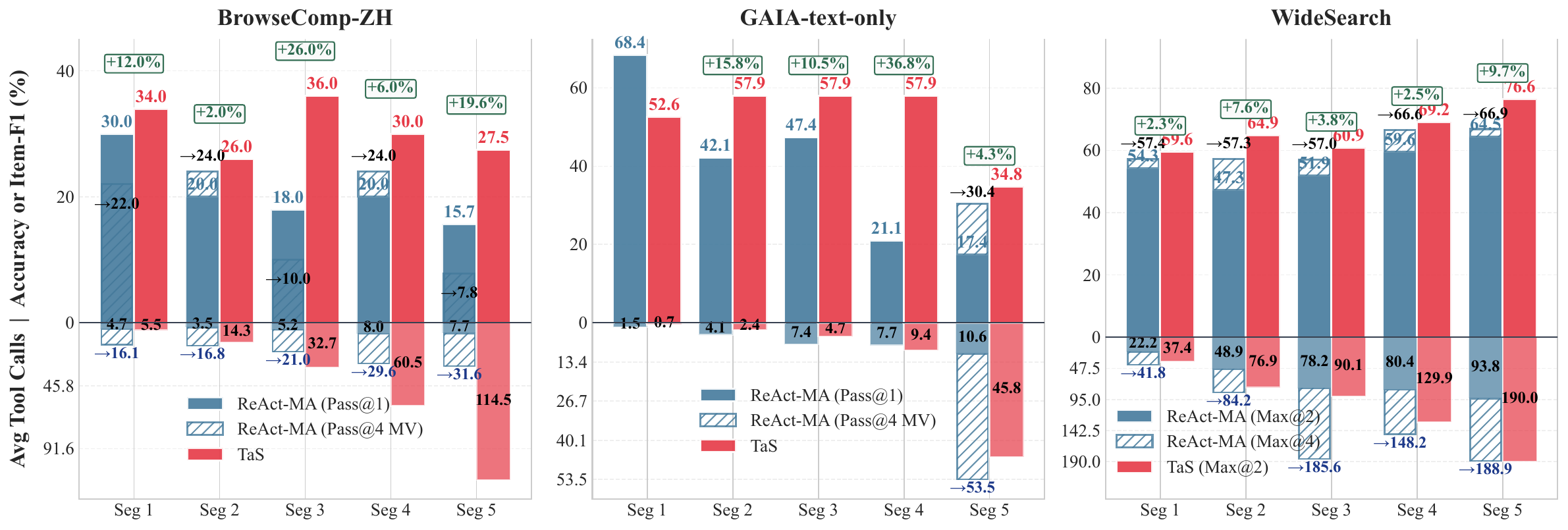}
    \caption{Search efficiency analysis of \texttt{Gemini-2.5-Flash} on Deep Search and Wide Search benchmarks.}
    \label{fig:main_architecture}
    %\vspace{-10pt}
\end{figure*}

\subsection{Search and Exploration Efficiency~\label{sec:analysis_cost_effectiveness}}

\paragraph{RQ2: Is performance driven by planning quality or strictly by search volume?}
To fairly test performance with comparable search efficiency, we categorize instances into five segments based on the number of tool usage (sorted by tool usage volume) and benchmark TaS against compute-scaled baselines:
ReAct-MA with Majority Voting (MV, $N$=4) for Deep Search, and ReAct-MA (Max@4) for Wide Search. As shown in Figure~\ref{fig:main_architecture}, TaS demonstrates qualitative superiority over scaling variant of baselines:
(1) \textbf{Deep Search}: For example, in the most demanding segment (Seg 5) of GAIA, TaS outperforms the ReAct-MA MV (+4.3\% improvement) while strictly consuming fewer tool calls (Avg. 45.8 $<$ 53.5), proving that superior search efficiency of TaS;
(2) \textbf{Wide Search}: Similarly, TaS (Max@2) significantly outperforms ReAct-MA (Max@4) across all segments, while TaS' tool usage is comparable or even less. This confirms that TaS's advantage stems from precise and effective structured planning and state management, not merely increased search volume.

Moreover, TaS ensures precise exploration of the search space in WideSearch, as measured by Num@k (i.e., the maximum valid cells, defined as $N_{total} \times \text{Item-P}$, achieved across $k$ trials). 
Table~\ref{tab:wide_search_table_size} shows that TaS Num@1 already surpasses ReAct-MA Num@4 (199.7 $>$ 199.4). Besides, TaS Num@4 closely approaches the ground truth upper bound (251.1 vs. 274.8). 

\begin{table}[htbp]
\centering
\setlength{\tabcolsep}{3pt}
\renewcommand{\arraystretch}{1.15}
\resizebox{\linewidth}{!}{
\begin{tabular}{l ccccc}
\toprule
\textbf{Method} & \textbf{Num@1} & \textbf{Num@2} & \textbf{Num@3} & \textbf{Num@4} & \textbf{GT} \\ \midrule
ReAct-SA & 139.3 & 159.0 & 169.3 & 172.6 &  \\
ReAct-MA & 158.0 & 186.0 & 194.7 & 199.4 &  \\
\rowcolor{lightgray}
TaS (Ours) & \textbf{199.7} & \textbf{211.4} & \textbf{229.4} & \textbf{251.1} & \cellcolor{white}\multirow{-3}{*}{274.8} \\
\bottomrule
\end{tabular}
}
\caption{Comparison on Num@k of \texttt{Claude-Sonnet-4}. GT denotes the upper bound in ground-truth tables.}
\label{tab:wide_search_table_size}
\vspace{-10pt}
\end{table}

\subsection{Test-time Scaling Analysis~\label{sec:analysis_testtimescaling}}
\begin{figure}[h]
    \centering
    \includegraphics[width=0.48\textwidth]{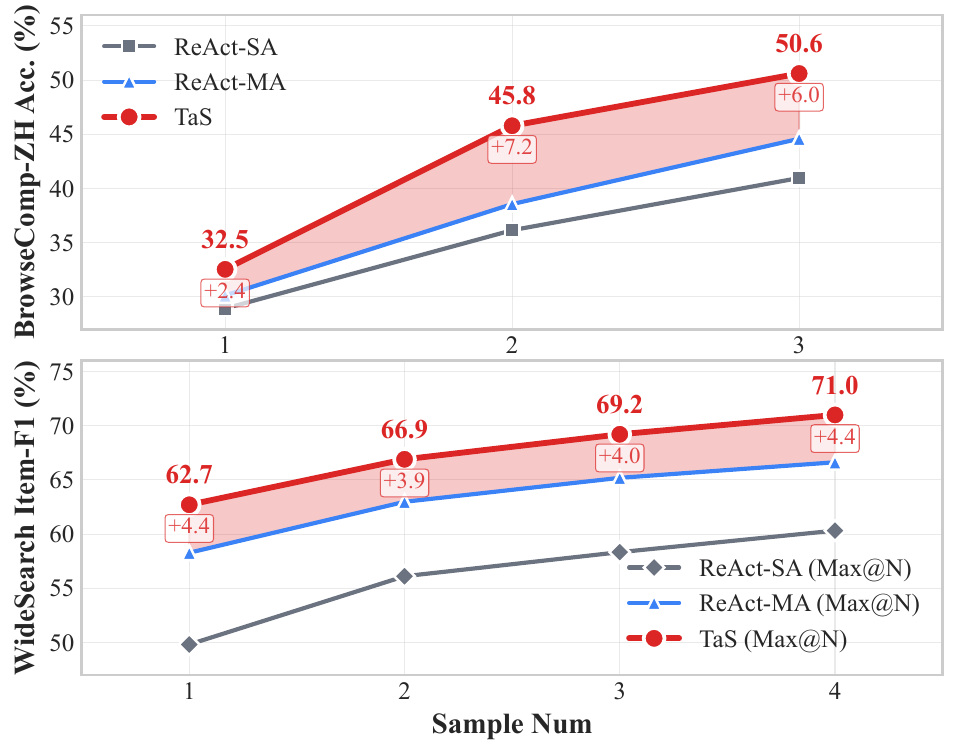}
    \caption{Test-time Scaling Analysis on BrowseComp-ZH (Top, \texttt{Gemini-2.5-Flash}) and WideSearch (Bottom, \texttt{Claude-Sonnet-4}).}
    \label{fig:tts_analysis}
    \vspace{-10pt}
\end{figure}

\paragraph{RQ3: Does the structured planner drive more effective exploration during test-time scaling?} 
We investigate whether allocating more inference compute benefits TaS more effectively than unstructured ReAct. Figure~\ref{fig:tts_analysis} illustrates the scaling trends on BrowseComp-ZH (Pass@N) and WideSearch (Max@N).
It can be observed that as the compute budget ($N$) expands, the performance gap widens. For instance, on BrowseComp-ZH, the performance gap between TaS and ReAct-MA widens from +2.4\% ($N$=1) to +7.2\% ($N$=2). On WideSearch, the advantage of TaS amplifies from +4.0\% ($N$=3) to +4.4\% ($N$=4). 
Besides, TaS at $N$=2 consistently exceeds ReAct-MA at $N$=3 (Deep Search) and $N$=4 (Wide Search). 
This demonstrates that TaS benefits more effectively from test-time scaling.

\subsection{Ablation Study on TaS Component~\label{sec:analysis_ablation}}
\paragraph{RQ4: Which component is the most critical: Planner Main-Agent or Sub-Agent?} 
Table~\ref{tab:ablation_study} reveals that the Planner Main-Agent is the critical bottleneck in our proposed framework: Downgrading the Planner from Qwen3-Max to Qwen3-30B-A3B causes a significant drop, while downgrading the Sub-Agent has a much milder impact.

Similar to findings in Section~\ref{sec:exp_deepwide_search}, TaS exhibits flexibility: As shown in the last four rows in Table~\ref{tab:ablation_study}, replacing the general \texttt{Gemini-2.5-Flash} Sub-Agent with the MiroThinker-8B deep search model~\cite{miromind2025mirothinker} yields a substantial performance improvement across most metrics. This indicates that the Sub-Agent is ``plug-and-play'', allowing specialized and cost-efficient models to replace larger foundation models.

More importantly, integrating MiroThinker-8B into TaS (w/ \texttt{Gemini}) significantly outperforms the standalone \texttt{MiroThinker-8B} model on all metrics. This validates that TaS effectively unlocks and amplifies the potential of specialized deep search models, proving the effectiveness of planner in TaS.

\begin{table}[h]
\centering
\renewcommand{\arraystretch}{1.2} %稍微增加行高，让表格看起来不拥挤
\resizebox{\linewidth}{!}{
\begin{tabular}{lcccc}
\toprule
\multirow{2}{*}{\textbf{Model Variant}} & \textbf{DeepSearch} & \multicolumn{3}{c}{\textbf{WideSearch}} \\
\cmidrule(lr){2-2} \cmidrule(lr){3-5}
 & \textbf{BC-ZH} & \textbf{Row-F1} & \textbf{Item-F1} & \textbf{Col-F1} \\
\midrule
\multicolumn{5}{c}{\cellcolor{gray!10}\textbf{TaS Framework: \texttt{Qwen3-235B-A22B} Sub-Agents. Planners are}} \\\midrule
+ Qwen3-Max       & \textbf{36.5} & \textbf{25.5} & \textbf{48.0} & 58.1 \\
+ Qwen3-235B & 29.9 & 14.6 & 36.5 & 50.6 \\
+ Qwen3-30B   & 7.1  & 8.6  & 22.9 & 33.3 \\
$\Delta$ (Qwen3-30B)      & \textcolor{red}{$\boldsymbol{\downarrow}$\textbf{29.4\%}} & \textcolor{red}{$\downarrow$16.9\%} & \textcolor{red}{$\boldsymbol{\downarrow}$\textbf{25.1\%}} & \textcolor{red}{$\boldsymbol{\downarrow}$\textbf{24.8\%}} \\

\midrule

\multicolumn{5}{c}{\cellcolor{gray!10}\textbf{TaS Framework: \texttt{Qwen3-Max} Planner. Sub-Agents are}} \\ \midrule
+ Qwen3-Max       & \textbf{38.0} & \textbf{38.5} & \textbf{57.8} & \textbf{66.9} \\
+ Qwen3-235B & 36.5 & 25.5 & 48.0 & 58.1 \\
+ Qwen3-30B   & 27.0 & 16.9 & 45.0 & 63.6 \\
$\Delta$ (Qwen3-30B) & \textcolor{red}{$\downarrow$11.0\%} & \textcolor{red}{$\boldsymbol{\downarrow}$\textbf{21.6\%}} & \textcolor{red}{$\downarrow$12.8\%} & \textcolor{red}{$\downarrow$3.3\%} \\ 
\midrule
\multicolumn{5}{c}{\cellcolor{gray!10}\textbf{TaS Framework: \texttt{Gemini-2.5-Flash} Planner. Sub-Agents are}} \\ \midrule
% /Users/qihui/Desktop/DeepWideSearch/Tabular-center-Agent/Exp/已有数据分析/20251217/outputs/output_deepsearch/browsecomp-zh/gemini-2.5-flash-tts-subset-test-v1
% /Users/qihui/Desktop/DeepWideSearch/Tabular-center-Agent/Exp/已有数据分析/20251217/outputs/output_widesearch/gemini-2.5-flash-tts-v3
+ Gemini-2.5-Flash & 33.0 & \textbf{32.7} & 52.5 & 65.8\\
+ MiroThinker-8B & \textbf{40.0} & 32.1 & \textbf{59.0} & \textbf{75.9} \\
\midrule \multicolumn{5}{c}{Compare with \textit{\texttt{MiroThinker-v1.0-8B}} Standalone Baseline} \\ \midrule
\small{Only MiroThinker} & 32.0 & 19.8 & 36.0 & 47.4 \\
$\Delta$ \small{(Only MiroThinker)} & \textcolor{red}{$\boldsymbol{\downarrow}$8.0\%} & \textcolor{red}{$\boldsymbol{\downarrow}$12.3\%} & \textcolor{red}{$\boldsymbol{\downarrow}$23.0\%} & \textcolor{red}{$\boldsymbol{\downarrow}$28.5\%}\\
\bottomrule
\end{tabular}
}
\caption{Ablation study on the subsets of two benchmarks. The row $(\Delta)$ indicates the performance drop.~\label{tab:ablation_study}}
\vspace{-10pt}
\end{table}

%% file: section/analysis_case_study.tex
\begin{figure*}[t]
\centering
% 开始 Case Study 盒子
\begin{casestudybox}{Case Study on BrowseComp-ZH \#141}
    % --- Query 部分 ---
    \noindent\textbf{Query:} \textit{Find a singer who graduated from a university in their birth province. Around age 20, they sang a theme song for a TV drama with two identical characters in the title. They released their first album around age 22. Who is this singer?}
    \vspace{0.5em}\hrule\vspace{0.5em}

    % --- Method A: ReAct Baseline ---
    \caseheader{ReAct Baseline}

    \noindent\textbf{Trajectory:}
    \begin{enumerate}[nosep, leftmargin=1.5em, label=\arabic*.]
        \item \texttt{Search}[Singer 20 years old TV theme song two identical characters...]
        \item \texttt{Search}[Zhao Wei birth place graduation first album] $\rightarrow$ \textit{Discarded}
        \item \texttt{Search}[Chinese singer 20 years old theme song 22 years old first album...]
        \item \texttt{Search}[Hu Xia TV theme song repeated characters]
        \item \texttt{Visit}[.../item/Hu\_Xia/...]
        \item \textbf{Conclusion:} \textcolor{red}{\xmark \textbf{Hu Xia (Incorrect)}}
    \end{enumerate}

    \vspace{0.4em}
    \noindent\textbf{Failure Analysis:} The model found a partial match (Hu Xia) and halted prematurely. It failed to explicitly verify the "First Album Age" constraint (Hu released his first album at age 20, not 22), leading to a false positive.

    % --- Method B: Ours ---
    \caseheader{Ours Proposed TaS Framework}

    \noindent\textbf{Process Overview:}
    \begin{itemize}[nosep, leftmargin=1.5em]
        \item \textbf{1. Schema Definition:} Columns defined for \textit{Birth Prov., Univ. Prov., Theme Songs (Age $\approx$ 20), Album Year (Age $\approx$ 22)}.
        \item \textbf{2. Search:} Retrieve 10 candidates (including \textit{Jiang Dunhao, Chen Lin, Liu Xijun, Hu Xia, Shan Yichun...}).
    \end{itemize}

    \vspace{0.6em}
    \noindent\textbf{3. Table Completion:}
    % --- 核心表格嵌入在这里 ---
    \begin{center}
    \small % 表格内容稍微缩小一点以适应盒子
    \setlength{\tabcolsep}{4pt} % 微调列间距
    % 使用 tabular* 确保表格宽度合适，或者用固定宽度的 p 列
    \begin{tabular}{@{} l c c p{0.43\linewidth} c @{}}
    \toprule
    \textbf{Candidate} & \textbf{Birth/Univ.} & \textbf{Match?} & \textbf{Theme Song (Age $\approx$ 20) \& Album (Age $\approx$ 22)} & \textbf{Verdict} \\
    \midrule
    Hu Xia & GX / GX & \cmark & Song: \textit{Summer Solstice} (Age 27) \newline Album: \textit{Hu Aixia} (2010, Age 20 $\neq$ 22) & \xmark \\
    \cmidrule{1-5}
    \textbf{Shan Yichun} & \textbf{ZJ / ZJ} & \cmark & \textbf{Song:} \textit{Xu Xie} (Drama: \textit{Yi Sheng Yi Shi}, 2021, Age 20) \newline \textbf{Album:} \textit{Brave Quota} (2022, Age 21\textsuperscript{*}) & \cmark \\
    \cmidrule{1-5}
    Liu Xijun & GD / GD & \cmark & Song: \textit{Bei Ke Feng Ling} (Age 18) \newline Album: \textit{Love Garden} (2010, Age 22) \newline \textit{\scriptsize *Fails on song title constraint} & \xmark \\
    ... & ... & ... & ... & ... \\
    \bottomrule
    \multicolumn{5}{l}{\scriptsize \textit{*Note: Age 21 is considered "around age 22" by the ground truth standard.}}
    \end{tabular}
    \end{center}
    % -------------------------

    \vspace{0.4em}
    \noindent\textbf{Correction Analysis:} By explicitly filling the schema, the agent identified that only \textcolor{blue}{\cmark \textbf{Shan Yichun (Correct)}} satisfied all constraints robustly, filtering out false positives like Hu Xia based on precise data points.

\end{casestudybox}
% Caption 和 Label 放在盒子外面
\caption{Case study between the ReAct and our proposed TaS Framework on the BrowseComp-ZH benchmark.}
\label{tab:case_study_deepsearch}
\end{figure*}

\begin{figure*}[t]
\centering
\begin{casestudybox}{Case Study on WideSearch \#EN-059}
    % --- Query 部分 (翻译为英文) ---
    \noindent\textbf{User Query:} \textit{Verify basic information for all TED Prize winners from 2005 to 2015. Required columns: [Year, Winner, TED Talk Title, Host City]. Output a Markdown table. Do not omit any cells; use "NA" if not found.}
    \vspace{0.5em}\hrule\vspace{0.5em}

    % --- Method A: Baseline ---
    \caseheader{Multi-Agent ReAct Baseline}

    \vspace{0.4em}
    \noindent\textbf{Outcome (Low Recall):}
    \begin{center}
    \small
    \begin{tabular}{@{} l l l l @{}}
    \toprule
    \textbf{Year} & \textbf{Winner} & \textbf{Talk Title} & \textbf{City} \\
    \midrule
    2005 & Bono & \textcolor{red}{NA} & \textcolor{red}{NA} \\
    2006 & Lawrence Brilliant & \textit{...} & Monterey \\
    2007 & Bill Clinton & \textcolor{red}{NA} & \textcolor{red}{NA} \\
    ... & ... & ... & ... \\
    2015 & Dave Isay & \textit{Everyone around ...} & Vancouver \\
    \bottomrule
    \end{tabular}
    \end{center}
    \vspace{0.2em}
    \noindent\textbf{Failure Analysis:} Without a global schema to track progress, the agent \textbf{lost context} during the multi-hop reasoning. It inadvertently \textbf{omitted} the search for the critical "First Album Age" constraint, jumping directly to an erroneous conclusion based on incomplete evidence.

    % --- Method B: Ours ---
    \caseheader{Our Proposed TaS Framework}

    \noindent\textbf{Process Overview:}
    \begin{itemize}[nosep, leftmargin=1.5em]
        \item \textbf{1. Schema \& Strategy:} Schema defined as \texttt{[Year, Winner, Title, City]}. The planner explicitly decomposes the time range: \textit{"Search 2005-2010 winners" and "Search 2011-2015 winners"}.
        \item \textbf{2. Row Expansion:} Parallel agents successfully retrieve all 11 winners (Rows) by cross-referencing multiple sources.
        \item \textbf{3. Cell Completion:} The planner detects missing "City" and "Title" values in the initial draft. Sub-agents are dispatched: e.g., \texttt{Search[Sylvia Earle TED Prize 2009 host city]}.
    \end{itemize}

    \vspace{0.4em}
    \noindent\textbf{Final Outcome (100\% Coverage):}
    \begin{center}
    \small
    \setlength{\tabcolsep}{3pt}
    \begin{tabular}{@{} l l p{0.35\linewidth} l @{}}
    \toprule
    \textbf{Year} & \textbf{Winner} & \textbf{Talk Title} & \textbf{City} \\
    \midrule
    2005 & Bono & \textit{Three unusual ...} & Monterey \cmark \\
    ... & ... & ... & ... \\
    2006 & Cameron Sinclair & \textit{A call for...} & Monterey \cmark \\
    ... & ... & ... & ... \\
    2007 & Bill Clinton & \textit{Rebuilding Rwanda} & Monterey \cmark \\
    ... & ... & ... & ... \\
    2009 & Sylvia Earle & \textit{Protect our oceans} & Long Beach \cmark \\
    ... & ... & ... & ... \\
    2015 & Dave Isay & \textit{Everyone around you...} & Vancouver \cmark \\
    \bottomrule
    \end{tabular}
    \end{center}
    \vspace{0.2em}
    \noindent\textbf{Conclusion:} By structuring the search horizon and employing targeted cell-filling, our method achieves \textbf{11/11 recall} for rows and completes all attribute columns, whereas the baseline suffers from significant omission.

\end{casestudybox}
\caption{Case Study on WideSearch Benchmark (Task \#EN-059). }
\label{tab:case_study_widesearch}
\end{figure*}

\begin{figure*}[t]
\centering
\begin{casestudybox}{Case Study on DeepWide Search (Task: US Lighting Merchants)}
    % --- Query 部分 ---
    \noindent\textbf{User Query:} \textit{Find 20 local US-based lighting manufacturers/merchants that operate on e-commerce platforms (Amazon, Walmart) or independent sites. Required columns: [Platform, Store Name, Email, Phone, Product Count].}
    \vspace{0.5em}\hrule\vspace{0.5em}

    % --- Method A: Baseline ---
    \caseheader{Multi-Agent ReAct Baseline}

    \noindent\textbf{Outcome (Incomplete \& Low Precision):}
    \begin{center}
    \small
    \setlength{\tabcolsep}{3pt}
    \begin{tabular}{@{} l l l l l l @{}}
    \toprule
    \textbf{Merchant} & \textbf{Platform} & \textbf{Store Name} & \textbf{Email} & \textbf{Phone} & \textbf{Verdict} \\
    \midrule
    Progressive Lighting & Amazon/Site & Lights Online & \textcolor{red}{NA} & (866) 688-3562 & \cmark \\
    Brand Name Lighting & Amazon & \textit{Generic Store} & \textcolor{red}{NA} & \textcolor{red}{NA} & \cmark \\
    \textbf{AvitaLights} & Etsy & David Avital & \textcolor{red}{NA} & \textcolor{red}{NA} & \xmark \textit{(Not US)} \\
    \textbf{HANM} & Etsy & \textcolor{red}{NA} & \textcolor{red}{NA} & \textcolor{red}{NA} & \xmark \textit{(Invalid)} \\
    Wholesale Lighting & Amazon & \textcolor{red}{NA} & \textcolor{red}{NA} & \textcolor{red}{NA} & \xmark \textit{(Not Found)} \\
    ... & ... & ... & ... & ... & ... \\
    \bottomrule
    \end{tabular}
    \end{center}
    \vspace{0.2em}
    \noindent\textbf{Performance:} \textbf{Column-F1: 80.0\%} (False positives), \textbf{Item-P: 73.0\%} (Missing contacts).
    \vspace{0.3em}
    \newline
    \noindent\textbf{Failure Analysis:} The baseline struggles with the \textbf{dual complexity} of breadth and depth. It fills slots with ineligible candidates (e.g., non-US Etsy sellers) to meet the "20 merchants" count and frequently fails to navigate to "Contact Us" pages for deep information extraction, resulting in empty email/phone cells.

    % --- Method B: Ours ---
    \caseheader{Our Proposed TaS Framework}

    \noindent\textbf{Process Overview:}
    \begin{itemize}[nosep, leftmargin=1.5em]
        \item \textbf{1. Wide Search (Filtered Expansion):} Parallel sub-agents scan Amazon/Google, filtering out non-US sellers like \textit{AvitaLights}.
        \item \textbf{2. Deep Search (Deep Crawling):} Targeted sub-agents visit official sites (e.g., \textit{meyda.com}, \textit{studio.hammerton.com}) to specifically locate contact details.
    \end{itemize}

    \vspace{0.4em}
    \noindent\textbf{Final Outcome (High Recall \& Precision):}
    \begin{center}
    \small
    \setlength{\tabcolsep}{2pt}
    % 使用 p 列来自动换行，防止表格过宽
    \begin{tabular}{@{} p{0.18\linewidth} p{0.1\linewidth} p{0.17\linewidth} p{0.18\linewidth} p{0.15\linewidth} p{0.1\linewidth} @{}}
    \toprule
    \textbf{Merchant} & \textbf{Platform} & \textbf{Store Name} & \textbf{Email} & \textbf{Phone} & \textbf{Verdict} \\
    \midrule
    Meyda Lighting & Site & Meyda.com & sales@meyda.com & 800-222-4009& \cmark \\
    LFI Lights & Amazon & Light Fixture Ind. & info@lightfixture... & 877-534-4621& \cmark  \\
    Hammerton Studio & Site & Hammerton Studio & info@studio.ham... & 801-973-8095& \cmark  \\
    HitLights & Amazon & HitLights & customer@hitli... & (855) 768-4135& \cmark  \\
    Commercial LED & Site & U.S. Wholesale & info@commercial... & (313) 528-7900 &\cmark  \\
    LightArt & Site & LightArt & info@lightart.com & 206-524-2223& \cmark  \\
    ... & ... & ... & ... & ... & ...\\
    \textbf{TorchStar} & Site & TorchStar & info@torchstar.us & (800) 990-7688& \cmark  \\
    \bottomrule
    \end{tabular}
    \end{center}
    \vspace{0.2em}
    \noindent\textbf{Performance:} \textbf{Column-F1: 95.0\%} (Correctly identified US merchants), \textbf{Item-P: 78.9\%} (Rich contact details).
    \vspace{0.3em}
    \newline
    \noindent\textbf{Conclusion:} In DeepWide tasks, our framework excels by first strictly verifying candidate eligibility (US-based) during row expansion, and then leveraging deep search capabilities to retrieve hard-to-find attributes (Emails/Phones), significantly outperforming the baseline in both entity quality and information density.

\end{casestudybox}
\caption{Case Study in our curated DeepWide Search Benchmark.}
\label{fig:case_study_deepwide}
\end{figure*}

%% file: section/appendix.tex
\newpage
\section{Experimental Details~\label{appendix:exp_setup}}

\subsection{Benchmarks and Metrics}

\paragraph{Deep Search Benchmark.} 
We employ the standard LLM-as-a-Judge evaluation protocol from BrowseComp-ZH~\cite{zhou2025browsecompzhbenchmarkingwebbrowsing} to assess the correctness of generated answers on both GAIA~\cite{mialon2023gaiabenchmarkgeneralai} and BrowseComp-ZH benchmarks.
For the ablation studies and efficiency analyses presented in Section~\ref{sec:analysis}, due to the high computational cost and API quota limitations, we utilize a representative subset of the BrowseComp-ZH dataset consisting of 100 randomly sampled instances.

\paragraph{Wide Search Benchmark.} 
We adopt the official evaluation framework of the WideSearch benchmark~\cite{wong2025widesearchbenchmarkingagenticbroad} to reproduce the ReAct baselines and compute standard metrics, including Row-F1, Item-F1, and Success Rate.
In addition to these metrics, we introduce a \textbf{Column-F1} metric to explicitly measure the accuracy of the retrieved entities within the table. This metric allows us to decouple the quality of entity discovery from the quality of information extraction.
Similar to the Deep Search setting, experiments in Section~\ref{sec:analysis} are conducted on a stratified subset of the WideSearch dataset containing 50 samples.

\paragraph{DeepWide Search Benchmark.}
% 当前社区缺乏公开的深宽搜索的数据集，这类数据集极其难以构建和评测。我们 follow 当前已有的技术实现方案，在我们关注的招商 BD 业务场景上构建了 20 条专业且能反应实际情况的数据集。
% 具体的一条深宽搜索数据如下tcolorbox所示，可以发现这类数据相比深度搜索和宽度搜索都要困难，其包含大量的限制条件信息，需要通过海量的搜索并从中进行深度搜索以筛选出符合要求的样本，同时还要对每个符合要求的样本的 information 信息（如联系人邮箱、官网等）进行大量补充收集。
% 样本 case：请帮我找以下公司的商务合作邮箱、商务合作电话：请给我 30 个面向西班牙市场、售卖 addidas 运动鞋、且有价格竞争力、有 B2C 运营经验的商家的如下联系方式（电话、合作邮箱、销售平台、公司官网、CEO 名称）。
%ground-truth 通过大量的人工专家标注、 agent 和闭源系统构建后，再由人工专家统一标注和确认。为了确保评测的可靠性，模型生成结果由至少 4 个人类专家标注并报告两类重要的指标：（1）Column-F1：评估检索到的样本的准确性；（2）Item-P：针对检索到的正确样本，评估其相关检索到的单元格信息的准确率。
The current research community lacks open benchmarks that simultaneously demand extensive horizontal breadth (identifying numerous entities) and vertical depth (complex constraints and attribute extraction). Such datasets are notoriously difficult to construct and evaluate.
To address this gap, we follow ~\citet{parallel2025findall} to create a specialized DeepWide dataset. This dataset consists of 20 high-quality, complex samples focused on Business Development (BD) scenarios, which reflect real-world industrial workflows.
As illustrated in Sample~\ref{box:deepwide_case}, DeepWide Search presents significantly higher complexity than isolated Deep or Wide search tasks. They require a rigorous two-stage process:
(1) Complex Filtering: The agent must scan massive amounts of information to identify entities that satisfy multiple strict constraints (e.g., target market, product category, pricing strategy);
(2) Deep Information Collection: For each identified entity, the agent must perform deep searches to retrieve specific missing details (e.g., contact emails, executive names).

Given the inherently open-ended nature of these tasks, constructing an exhaustive ground-truth universe is computationally infeasible. To ensure robust yet manageable evaluation, we implemented a strict protocol: First, we explicitly constrain the retrieval target to a fixed quantity for each query (e.g., 30 candidates) to bound the search space.
we construct the ground truth via a dynamic union strategy, aggregating verified correct matches from commercial state-of-the-art systems (like EXA.ai and Gemini DeepResearch etc.), our internal baselines (ReAct-MA and TaS), and expert annotation.
To ensure reliability, the final ground truth was unified and verified by domain experts.
This reference dataset is rigorously verified by domain experts, who also maintain an exclusion list for known false positives. This dynamic mechanism allows us to iteratively update the ground truth table, significantly reducing annotation costs while ensuring high-fidelity assessment for future evaluations.

Given the open-ended nature of these tasks, the experimental results are annotated by four experts engaged in business development (BD) applications, each holding at least a master’s degree.
The hourly wage of our human annotators is over \$34, which
is much higher than average hourly wage \$3.13 on Amazon Mechanical Turk~\citep{hara2017datadrivenanalysisworkersearnings}.
We report two primary metrics for this benchmark: (1) \textbf{Column-F1:} Evaluates the accuracy of the identified entities against the complex constraints; (2) \textbf{Item-Precision (Item-P):} Measures the accuracy of the retrieved information specifically for the correctly identified entities.

%%%% DeepWide Case Box %%%%
\begin{tcolorbox}[
    colback=gray!5,       % 背景色，可改为你的自定义颜色如 boxback
    colframe=gray!60,     % 边框色，可改为你的自定义颜色如 boxframe
    title=\textbf{Sample~\ref{box:deepwide_case}: An Example of Our Curated DeepWide Search Benchmark},
    fonttitle=\small\sffamily,
    arc=2mm, boxrule=0.8pt,
    left=4pt, right=4pt, top=4pt, bottom=4pt
]
\small
\textbf{User Query:} 
Please help me identify \textbf{30 merchants} that meet all the following criteria:
\begin{enumerate}[itemsep=0pt, topsep=2pt, parsep=0pt, label=(\arabic*)]
    \item Target the \textbf{Spanish market};
    \item Sell \textbf{Adidas} sneakers;
    \item Offer \textbf{competitive pricing};
    \item Possess mature \textbf{B2C operational experience}.
\end{enumerate}
\vspace{2pt}
\textbf{Required Information:} 
For each identified merchant, retrieve the following contact details:
\textit{[Phone Number, Cooperation Email, Sales Platform, Official Website, CEO Name, Source URL]}.
\label{box:deepwide_case}
\end{tcolorbox}

\subsection{Fine-tuning Deep Search Sub-Agent~\label{appendix:finetune_32B}}

This section provides the details of our fine-tuned 32B model utilized in Section~\ref{sec:exp_deepwide_search}:

\paragraph{Base Model.} We utilized Qwen3-32B as the backbone for our Deep Search Sub-Agent. This 32B-parameter scale offers the optimal trade-off between reasoning capability and computational efficiency compared to smaller (14B) or larger variants (72B).

\paragraph{Data Construction.} We constructed a training dataset of approximately 12K samples using a hybrid strategy that combines trajectory distillation with reverse-synthesis to ensure diversity and robustness: (1) Trajectory Distillation (Forward): Following the trajectory collection paradigm of WebSailor~\cite{li2025websailornavigatingsuperhumanreasoning}, we collected multi-constraints user queries and distilled high-quality navigation trajectories. To ensure data quality, we implemented a rigorous iterative filtering pipeline. This involved removing unanswerable queries, employing a teacher LLM to parse and verify the format of search results, and optimizing the phrasing of questions based on ground-truth answers (hindsight relabeling). This yielded 11k high-quality samples; and (2) Reverse Synthesis (Reverse): To mitigate data sparsity for complex conditions, we employed a reverse-generation approach. We first sampled structured constraints to generate SQL queries and retrieve ground-truth candidates. These structured records were then converted into natural language templates and paraphrased into human-like complex search queries. This process contributed 1k samples specifically targeting multi-constraint reasoning.

\paragraph{Training Implementation.} The model is trained using Supervised Fine-Tuning (SFT) within 64k context windows. Learning rate is $5\times10^{-5}$ . The training is conducted on a computation cluster of 64 NVIDIA A100 GPUs within five hours.

\paragraph{Inference Settings.} We set the a maximum context window of 32B model as 64K. This extended context capability is critical for maintaining global coherence during deep search sessions, allowing the agent to process extensive search results and retain long-term history without truncation.

\subsection{Tools for Table Operation\label{appendix:table_operation}}

Our tabular memory system is built on MongoDB with PyMongo interfaces to ensure scalable and persistent state management. We expose six atomic primitives for agent interaction:

\begin{itemize}
    \item $\texttt{create\_table(schema)}$: Initializes the table structure based on the query-derived schema.
    \item $\texttt{add\_records(data)}$: Inserts new candidate entities (rows) discovered during the expansion phase.
    \item $\texttt{update\_records(filter, update)}$: Modifies specific cells to populate missing attributes for targeted candidates.
    \item $\texttt{show\_table(limit)}$: Serializes the current table snapshot into Markdown format for planner inspection.
    \item $\texttt{count\_table(filter)}$: Returns the number of rows matching specific criteria to verify target quantity.
    \item $\texttt{filter\_records(query)}$: Retrieves subsets of records (e.g., rows with empty cells) to isolate pending tasks.
\end{itemize}

All data manipulation operations (insertion, updates, and filtering) strictly adhere to standard PyMongo syntax (e.g., utilizing operators like \$set, \$exists). This enables the agent to perform precise logical queries natively within the database.

\subsection{Experimental Setup for Analysis\label{appendix:exp_setup_analysis}}

\paragraph{Computing Complexity.}
To rigorously evaluate model performance across varying degrees of task difficulty, we classify the samples in Deep Search (BrowseComp-ZH) and Wide Search (WideSearch) benchmarks into five distinct difficulty categorizations: \textit{Easy, Med-Easy, Medium, Med-Hard}, and \textit{Hard}. The specific complexity metrics for each benchmark are defined as follows: (1) \textbf{Deep Search:} We quantify complexity based on the number of search constraints within the user query. We utilized \texttt{Gemini-2.5-Flash} to parse each query and enumerate these constraints. A higher constraint count necessitates more intricate multi-hop reasoning and stricter information filtering, thereby increasing task difficulty; (2) \textbf{Wide Search:} We determine difficulty based on the size of the ground-truth table (the number of the table celss). Larger tables inherently demand a higher volume of search interactions to achieve full coverage, directly corresponding to a longer interaction horizon.

\paragraph{Experiments on Subset.}
Due to limited API quotas, test-time scaling and ablation study are conducted on the sampled subsets of 100 BrowseComp-ZH and 40 WideSearch samples.

\section{Detailed Process of TaS\label{appendix:show_case_control_tas}}

\begin{figure*}[h]
    \centering
    \includegraphics[width=\textwidth]{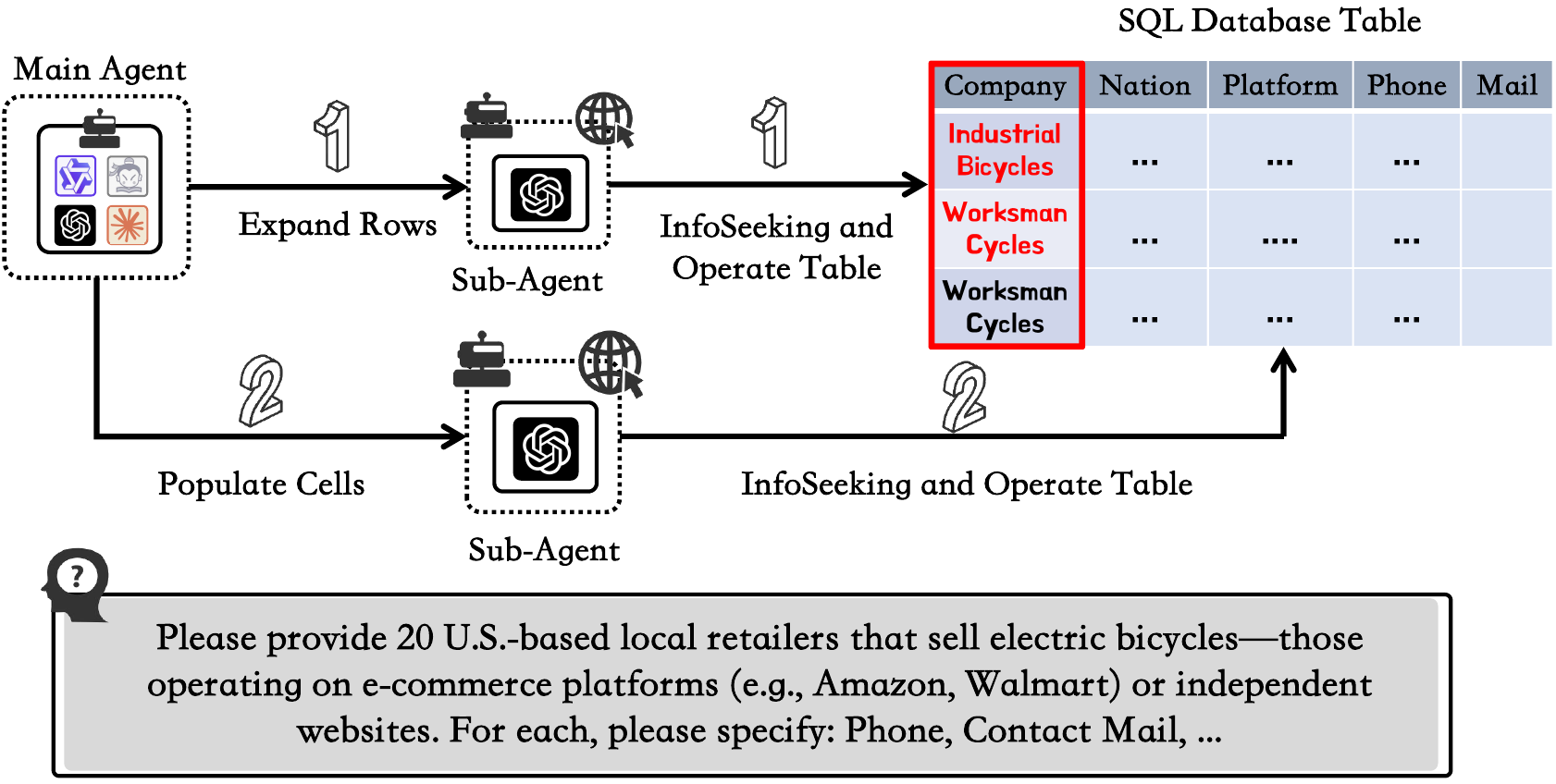}
    \caption{The detailed process of TaS on a complex DeepWide Search case in our benchmark.}
    \label{fig:show_case_control_tas}
\end{figure*}

The detailed process of our proposed TaS are shown in Figure~\ref{fig:show_case_control_tas}, aligning with the Algorithm~\ref{alg:table_execution}.

\section{More Experimental Results~\label{appendix:more_exp_rest}}

\subsection{Full Results on GAIA~\label{appendix:full_gaia_resukt}}
Table~\ref{tab:gaia_breakdown_full} provides the complete results of GPT-5, Qwen3-Max and Gemini-2.5-Flash on GAIA samples. It can be found that TaS consistently outperforms state-of-the-art baselines, while its performance is instable on tasks that do not require searching.

\begin{table}[htbp]
\centering
% Caption 说明：强调我们在 "Requires Search" 子集上的一致性优势
\resizebox{\linewidth}{!}{
\begin{tabular}{ll ccc}
\toprule
\textbf{Model} & \textbf{Sub-Task Type} & \textbf{ReAct} & \textbf{Ours} & \textbf{$\Delta$} \\
\midrule
% GPT-5 部分
\multirow{3}{*}{\shortstack[l]{\textbf{GPT-5}\\\textbf{Medium}\\\textbf{Think}}} 
 & Requires Search & 66.25\% & \textbf{71.25\%} & \textcolor{teal}{+5.0\%} \\
 & No Search & \textbf{91.30\%} & 86.96\% & \textcolor{red}{-4.34\%}\\
 \cmidrule(lr){2-5}
 & \textit{Overall} & 71.84\% & \textbf{77.67\%} & \textcolor{teal}{+5.87\%}\\
\midrule
% Qwen3-Max 部分
\multirow{3}{*}{\shortstack[l]{\textbf{Qwen3}\\\textbf{-Max}}} 
 & Requires Search & 46.84\% & \textbf{49.37\%} & \textcolor{teal}{+2.53\%} \\
 & No Search & \textbf{68.18\%} & 50.00\% & \textcolor{red}{-18.18\%} \\
 \cmidrule(lr){2-5}
 & \textit{Overall} & \textbf{51.49\%} & 49.50\% & \textcolor{red}{-1.98\%} \\
\midrule
% Gemini-2.5-Flash 部分
\multirow{3}{*}{\shortstack[l]{\textbf{Gemini}\\\textbf{2.5-Flash}}} 
 & Requires Search & 34.18\% & \textbf{49.37\%} & \textcolor{teal}{+15.19\%} \\
 & No Search & 55.00\% & \textbf{60.00\%} & \textcolor{teal}{+5.00\%} \\
 \cmidrule(lr){2-5}
 & \textit{Overall} & 38.38\% & \textbf{51.52\%} & \textcolor{teal}{+13.13\%} \\
\bottomrule
\end{tabular}
}
\caption{Detailed Performance on GAIA: samples requiring search or not ($N_r=80$ and $N_{nr}=23$).}
\label{tab:gaia_breakdown_full}
\end{table}

\begin{table}[H]
\centering
% 保持紧凑的列间距
\setlength{\tabcolsep}{3pt}
% 保持舒缓的行高
\renewcommand{\arraystretch}{1.15}
\resizebox{\linewidth}{!}{
% 修改这里：增加了一列 c (总共6列)
\begin{tabular}{l c cccc}
\toprule
\multirow{2}{*}{\textbf{Model}} & \textbf{ReAct} & \textbf{SR} & \textbf{Row} & \textbf{Item} & \textbf{Col} \\
& \textbf{Type} & \textbf{Acc} & \textbf{F1} & \textbf{F1} & \textbf{F1} \\
\midrule
% 修改这里：跨列数改为 6
\multicolumn{6}{c}{\textbf{Foundation Models with Tools}} \\
\midrule
Claude-S4 Think & SA & 5.0 & 41.9 & 66.7 & - \\
Claude-S4 Think & MA & 6.5 & 52.2 & 73.1 & - \\
Gemini-2.5-Pro  & SA & 5.0 & 41.4 & 63.6 & -\\
Gemini-2.5-Pro  & MA & 6.5 & 44.6 & 66.3 & -\\
OpenAI o3       & SA & 9.0 & 44.1 & 62.3 & - \\
OpenAI o3       & MA & 9.5 & 50.5 & 68.9 & - \\
KIMI-K2         & SA & 3.5 & 41.4 & 65.1 & - \\
KIMI-K2         & MA & 6.5 & 49.6 & 70.7 & - \\
\midrule
% 修改这里：跨列数改为 6
\multicolumn{6}{c}{\textbf{Our proposed TaS Framework}} \\
\midrule
Gemini-2.5-Flash & SA & \textbf{5.0} & 41.1 & 64.8 & 78.0 \\
Gemini-2.5-Flash & MA & 4.5 & 42.3 & 61.7 & 71.4  \\
\rowcolor{lightgray}
Gemini-2.5-Flash (Ours) & MA & \textbf{5.0} & \textbf{45.7} & \textbf{67.6} & \textbf{82.2} \\
Claude-S4 NoThink & SA & 4.5 & 38.1 & 60.9 & 74.1 \\
Claude-S4 NoThink & MA & 4.0 & 46.8 & 66.9 & 78.2 \\ 
\rowcolor{lightgray}
Claude-S4 NoThink (Ours) & MA & \textbf{9.1} & \textbf{49.0} & \textbf{71.0} & \textbf{84.4} \\
\bottomrule
\end{tabular}
}
\caption{\textbf{Max@4} Performance on WideSearch benchmark. Claude-S4 refers to \texttt{Claude-Sonnet-4}. SR denotes Success Rate. Results of baselines are copied from the paper~\cite{wong2025widesearchbenchmarkingagenticbroad}, where their Column-F1 scores are not recorded.}
\label{tab:wide_search_full_max@4}
\end{table}

\subsection{Max@4 Performance on WideSearch~\label{appendix:complete_widesearch_results}}
Beyond the stable Avg@4 metrics, we also analyze the Max@4 performance to assess the upper bound of agent capabilities in massive information aggregation. As detailed in Table~\ref{tab:wide_search_full_max@4}, TaS consistently unlocks superior potential compared to unstructured ReAct baselines. Most strikingly, TaS instantiated with the standard \texttt{Claude-Sonnet-4 (NoThink)} achieves a Success Rate of 9.1\%, significantly surpassing the computationally heavier Multi-Agent ReAct equipped with \texttt{Claude-Sonnet-4 (Thinking)} (6.5\%). This suggests that structured planning and state management is more critical than internal chain-of-thought reasoning for massive long-horizon search. Furthermore, this architectural advantage allows smaller models to punch above their weight. The lightweight \texttt{Gemini-2.5-Flash} with TaS outperforms the much stronger \texttt{Gemini-2.5-Pro (Multi-Agent ReAct)} across key metrics, achieving higher Row-F1 (45.7\% vs. 44.6\%) and Item-F1 (67.6\% vs. 66.3\%). This confirms that TaS effectively decouples performance from pure model scale, offering a cost-effective solution for industrial applications.

\subsection{Search and Exploration Efficiency~\label{appendix:cost_effectiveness}} 
\begin{figure}[H]
    \centering
    \includegraphics[width=0.48\textwidth]{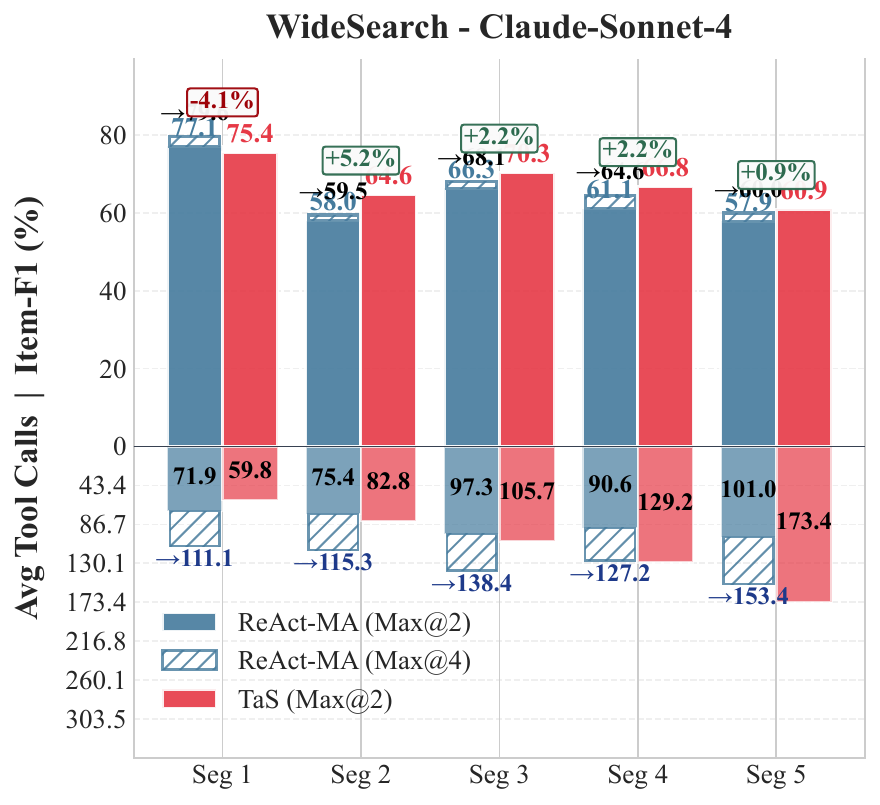}
    \caption{Search Efficiency Analysis on WideSearch of \texttt{Claude-Sonnet-4} model.}
    \label{fig:cost_efficiency_analysis_widesearch_claude}
\end{figure}
High performance in existing agents often comes at the cost of excessive interaction. However, Figure~\ref{fig:cost_efficiency_analysis_widesearch_claude} reveals that TaS breaks this trade-off. On the WideSearch benchmark, TaS (\texttt{Claude-Sonnet-4 (NoThink)}) attains these performance gains with comparable or even lower tool usage volume than the Multi-Agent ReAct baseline. This demonstrates that the performance gains stem from structured planning precision rather than brute-force search scaling.

\subsection{Robustness Analysis on WideSearch~\label{appendix:robustness_analysis}}

\begin{figure}[H]
    \centering
    \includegraphics[width=0.48\textwidth]{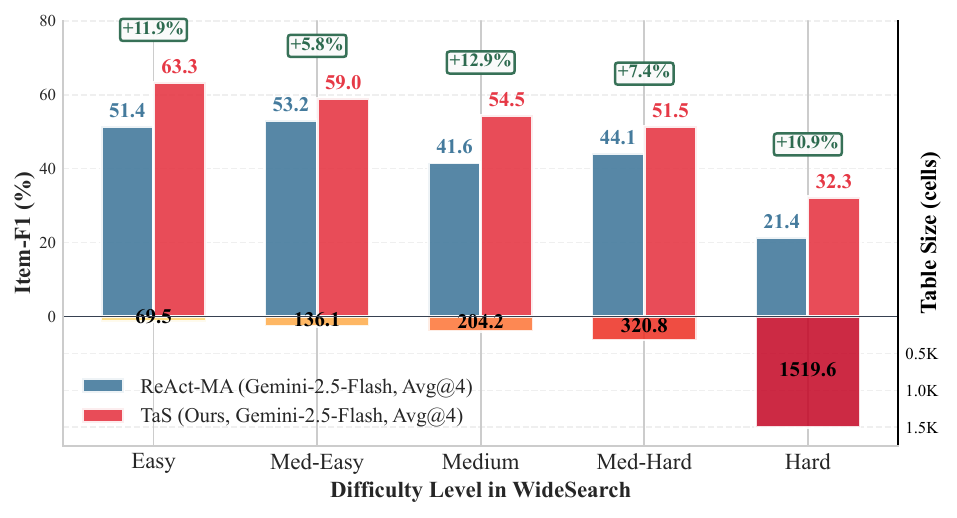}
    \caption{Search Efficiency Analysis on WideSearch of \texttt{Gemini-2.5-Flash} model.}
    \label{fig:cost_efficiency_analysis_widesearch_gemini}
\end{figure}

Figure~\ref{fig:cost_efficiency_analysis_widesearch_gemini} demonstrates that TaS consistently outperforms the Multi-Agent ReAct baseline across all difficulty tiers. The advantage is most critical in the "Hard" setting, where the state space explodes to over 1,500 cells. While the baseline collapses to 21.4\% Item-F1 under this cognitive load, TaS maintains robust performance at 32.3\% (+10.9\%). This confirms that structured planning effectively stabilizes small models against extreme context overload.

\section{Case Study~\label{appendix:case_study}}

\subsection{Qualitative Analysis~\label{appendix:case_study_comparison}}

Our case studies highlight how the table-centric design mitigates two critical failure modes of unstructured agents: (1) \textbf{Preventing Premature Convergence (Deep Search):} As shown in Figure~\ref{tab:case_study_deepsearch}, ReAct baselines often halt at partial matches (e.g., identifying "Hu Xia" but ignoring the album age). Our framework enforces \textbf{Global Verification} through schema filling, compelling the agent to validate every constraint against multiple candidates, thus filtering false positives; (2) \textbf{Eliminating Lazy Search (Wide Search):} As shown in Figure~\ref{tab:case_study_widesearch} and Figure~\ref{fig:case_study_deepwide}, baselines struggle with long-horizon retrieval, resulting in missing rows and empty cells. In contrast, our planner ensures Completeness by decomposing the search space (e.g., by year) for row expansion and dispatching targeted sub-agents for cell population.

\subsection{Search and No-Search Cases in GAIA~\label{appendix:case_study_gaia}}

To evaluate our framework's adaptability, we stratified the GAIA validation set based on the ground-truth tool usage annotations provided in the dataset metadata. We identified 80 search-dependent samples (where the solution requires web interaction) and 23 no-search samples (where the solution relies solely on internal reasoning, calculation, or coding).
Figure~\ref{fig:case_study_gaia_comparison} contrasts the distinct behavioral requirements of these two categories.

\begin{figure}[H]
\centering
\small
%\begin{casestudybox}{\small{Comparison on GAIA Samples}}
\begin{tcolorbox}[colback=gray!5, colframe=gray!60, arc=2mm, boxrule=0.8pt, title=\textbf{Comparison on GAIA Samples}, fonttitle=\small\sffamily, left=4pt, right=4pt, top=4pt, bottom=4pt]
\small
    
    % =============================================
    % CASE 1: No-Search Task (Logic/Math)
    % =============================================
    %\caseheader{No-Search Task}
    \textbf{No-Search Task}
    \noindent\textbf{Query:} \textit{Given the operation * defined on set $S = \{a, b, c, d, e\}$ via the table below. Provide the subset of $S$ involved in any counterexample proving * is not commutative. Format: alphabetical list.}
    \vspace{5pt}
    \hrule
    \vspace{5pt}
    % =============================================
    % CASE 2: Search Task (Information Retrieval)
    % =============================================
    %\caseheader{Search Task}
    \textbf{Search Task}
    \noindent\textbf{Query:} \textit{If all articles published by Nature in 2020 (articles only, not reviews) relied on statistical significance ($p=0.04$), how many papers would be incorrect? Round up.}

%\end{casestudybox}
\end{tcolorbox}
\caption{Comparative Analysis on GAIA.}
\label{fig:case_study_gaia_comparison}
\end{figure}

\section{The Use of Large Language Models}
In preparing this manuscript, Qwen-Max and Gemini 3 are used solely as a writing assistant to improve grammar and clarity. The LLMs was not used for generating code, concepts, or any part of the core research methodology.